%% file: main.tex
\documentclass{article}

\usepackage[preprint]{neurips_2025}

\usepackage[utf8]{inputenc} 
\usepackage[T1]{fontenc}    
\usepackage{hyperref}       
\usepackage{url}            
\usepackage{booktabs}       
\usepackage{amsfonts}       
\usepackage{natbib} 
\usepackage{textgreek}      
\usepackage{nicefrac}       
\usepackage{microtype}      
\usepackage{xcolor}         

\usepackage{amsmath,amssymb}
\usepackage{graphicx}
\usepackage{wrapfig}
\usepackage{multirow}

\title{Layerwise Recall and the Geometry of Interwoven Knowledge in LLMs}

\author{%
  Ge Lei \\
  Dyson School of Design Engineering\\
  Imperial College London\\
  SW72AZ, United Kingdom \\
  \texttt{g.lei23@imperial.ac.uk} \\
  \And
  Samuel J. Cooper \\
  Dyson School of Design Engineering \\
  Imperial College London\\
  SW72AZ, United Kingdom \\
  \texttt{samuel.cooper@imperial.ac.uk} \\
}

\begin{document}

\maketitle

\begin{abstract}

This study explores how large language models (LLMs) encode interwoven scientific knowledge, using chemical elements and LLaMA-series models as a case study. We identify a 3D spiral structure in the hidden states that aligns with the conceptual structure of the periodic table, suggesting that LLMs can reflect the geometric organization of scientific concepts learned from text. Linear probing reveals that middle layers encode continuous, overlapping attributes that enable indirect recall, while deeper layers sharpen categorical distinctions and incorporate linguistic context. These findings suggest that LLMs represent symbolic knowledge not as isolated facts, but as structured geometric manifolds that intertwine semantic information across layers. We hope this work inspires further exploration of how LLMs represent and reason about scientific knowledge, particularly in domains such as materials science.

\end{abstract}

\section{Introduction}

Large language models (LLMs) have demonstrated strong factual recall across a wide range of domains, from history and geography to science and mathematics \citep{naveed2023comprehensive, chang2024survey, kaddour2023challenges}. However, much remains unknown about how interconnected knowledge is internally organized. Clarifying these mechanisms is vital for aligning LLMs with human values \citep{ji2023ai}, enhancing their design, and broadening their applications.

Mechanistic interpretability offers a pathway to answer these questions \citep{bereska2024mechanistic, singh2024rethinking, dar2022analyzing}. Prior work suggests that models often represent features as linear directions in activation space \citep{gurnee2023language, tigges2023linear}. However, growing evidence reveals non-linear and geometrically structured patterns—such as cycles and modular circuits in tasks involving periodic concepts like days and months\citep{engels2024not}. These observations raise a deeper question: can LLMs represent interwoven knowledge in forms that mirror more complex conceptual geometry in the real world? 

Prior studies on factual recall have shown that core facts are enriched in early and middle multilayer perceptron (MLP) layers, propagated by attention, and linearly decodable via circuit-level analysis\citep{meng2022locating, geva2023dissecting, nanda2023factfinding}. They largely focus on isolated attributes, leaving open how multiple, interrelated facts are jointly encoded and retrieved.

Building on this, we delve into how LLMs encode and recall complex, interwoven knowledge through global, structured, and geometry-aware representations that evolve across the model’s depth. The contribution of our study are:

1. We report the first observation of a 3D spiral structure in LLM hidden states that organizes chemical elements in alignment with the structure of periodic table (Sec \ref{sec:geometric}).

2. We compare regression and classification probing, showing that middle layers encode continuous attribute structure, while later layers sharpen boundaries for fine-grained decisions (Sec \ref{sub:probing}).

3. We show that linguistic structure increasingly shapes knowledge representations in later layers (Sec \ref{sub:language}).

4. We find that LLMs recall related attributes through strong linear associations in middle layers, which weaken in deeper layers  (Sec \ref{sec:indirect_recall}).

\begin{wrapfigure}{r}{0.47\textwidth}
  \centering
  \includegraphics[width=1\linewidth]{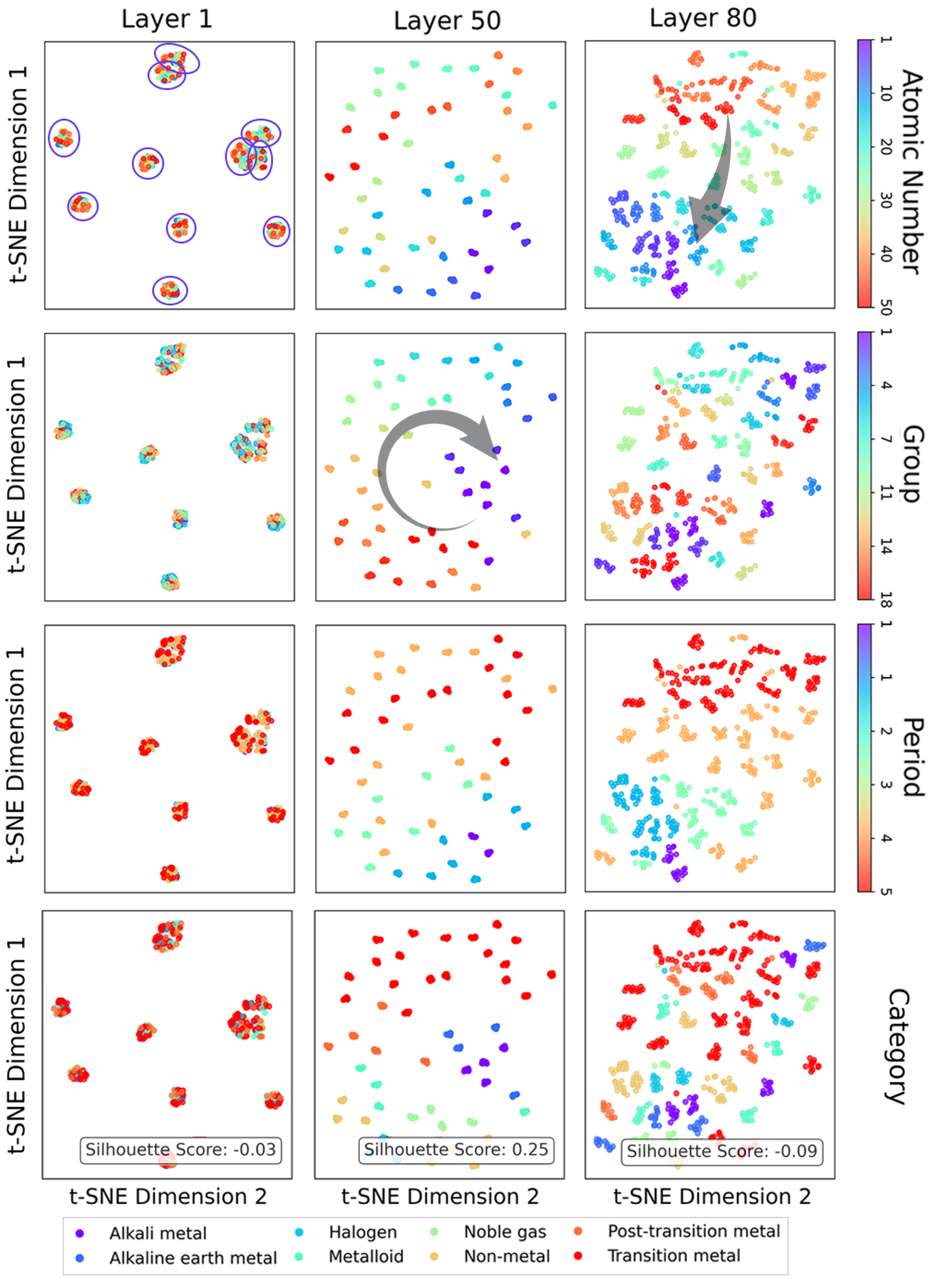}
  \caption{t-SNE visualization of Meta-Llama-3.1-70B last-token residual streams from the 1st, 50th, and 80th layers, using 11 continuation-style templates across 50 elements (550 points per plot). Each column shows one layer, while rows represent different colormaps highlighting attributes: `atomic number', `group', `period', and `category'. In the top-left plot, circled clusters correspond to individual templates, each containing 50 points.}
  \label{fig:vis}
\end{wrapfigure}

\section{Preliminaries}

Our study only focuses on how reliably acquired knowledge (\textit{i.e.} things we’re confident the model knows) is represented within LLMs, and excludes hallucinations or information not in the training set. We use the properties of chemical elements in the periodic table as a case study due to their frequent occurrence in training data, well-defined attributes, and quantifiable properties, making them an ideal subject for this investigation. We adopt Llama series models \citep{touvron2023llama, grattafiori2024llama} in this study.

\subsection{Residual stream collection}
To study how LLMs represent attributes across layers, we construct a prompt dataset based on a set of attributes (\(A = \{A_j\}_{j=1}^M\), such as `atomic number' or `group') and a set of elements (\(X = \{X_i\}_{i=1}^N\), such as `Mg' or `Al'). For linguistic diversity, we incorporate predefined template sets: \(T^{\text{cont}} = \{T_k^{\text{cont}}\}_{k=1}^{11}\) for continuation-style prompts and \(T^{\text{ques}} = \{T_k^{\text{ques}}\}_{k=1}^{11}\) for question-style prompts, with 11 templates in each.

In the continuation-style templates, the next output token would be the factual knowledge directly such as:  
\[
T_1^{\text{cont}}(A_j, X_i) = \text{`The } A_j \text{ of } X_i \text{ is '}
\]
\[
T_2^{\text{cont}}(A_j, X_i) = \text{`} X_i\text{'s } A_j \text{ is '}
\]

In question-style templates, the next output token is typically a syntactic word like `The', which ensures the grammatical structure is correct, such as:
\[
T_1^{\text{ques}}(A_j, X_i) = \text{`What is the } A_j \text{ of } X_i\text{?'}
\]
\[
T_2^{\text{ques}}(A_j, X_i) = \text{`Which value represents } X_i\text{'s } A_j\text{?'}
\]

By substituting each element and attribute (\(X_i\), \(A_j\)) into these templates, we generate prompts:  
\[
p_{i,j,k} = T_k(X_i, A_j)
\]

Each prompt \(p_{i,j,k}\) can then be fed into LLMs to study the corresponding residual streams at different layers.  

Last-token residual streams capture the full prompt context in decoder-only models with masked attention, as they integrate information from all preceding tokens. For each layer \(l\), we collect last-token residual streams \(\mathbf{h}_{i,j,k}^{(l)}\) from prompts \(p_{i,j,k}\) across all elements and templates (see Appendix~\ref{app:activation_collection} for details).

\subsection{Residual stream distribution}
\label{tsne}
We start with a preliminary visualization of the distribution of last-token residual streams for the `atomic number' attribute. Residual streams from each transformer layer \( l \) were collected for the atomic number attribute across the first 50 elements using 11 continuation-style templates, forming the set \(H_{\mathrm{Atomic Number}}^{(l)}\). To enable informative plots to be produced efficiently, PCA was applied to them and then t-SNE was use to project the first 50 principle components into 2D. Fig.\ref{fig:vis} shows the resulting distributions for Meta-Llama-3.1-70B, with points colored by atomic number as well as the other attributes (to visualize their association to atomic number).

The first column of the figure colors residual streams by true atomic number values (explicitly mentioned in the prompt). In early layers, prompts with similar vocabulary cluster together irrespective of atomic number, reflecting token-level similarity. In the middle layers, residual streams for the same element form small, heavily overlapping clusters, each containing representations from the 11 prompts generated for that element. By the final layers, although these points (same colors) still cluster together, individual points become distinguishable due to increased spread within the clusters.

In the next three columns, the residual streams are colored by the true values of attributes unmentioned in the prompt: `group', `period', and `category'. Despite not being mentioned in the prompt, in middle layers, chemically similar elements (small clusters with similar colors) cluster closely together. By the final layers, the clustering of some attributes, such as group and category, becomes less coherent, indicating a shift in representation. 

Furthermore, the geometric shape of attribute distributions varies. For example, atomic numbers form a linear arrangement transitioning from red to purple (1st row, 3rd column), while the `group' attribute activities form a cyclic pattern with sequential transitions (2nd row, 2nd column), potentially reflecting periodic relationships.


These visualizations suggest that LLM residual streams may encode attribute relationships in a structured and potentially geometric manner that reflect properties of the physical world. In particular, intermediate layers appear to capture implicit similarities even for unmentioned attributes, while later layers are tuned for task-specific outputs. These hypotheses are examined in the following sections.

\section{Geometric relationships among attributes}
\label{sec:geometric}
In materials science, a spiral trend emerges from the periodic variation in valence electron configurations as atomic number increases. By arranging elements sequentially and mapping their properties in a polar coordinate system, this periodicity becomes visually apparent as a spiral. We investigate whether LLMs, trained on extensive data about elemental properties, inherently capture these physical periodicities and reflect similar spiral structures in their learned embeddings.

We hypothesize that attributes in LLMs exist in a high-dimensional space, manifesting as linear, circular, or spiral patterns based on their structure, and then proceed to validate these geometries.

\begin{figure}
\centering
    \includegraphics[width=1\linewidth]{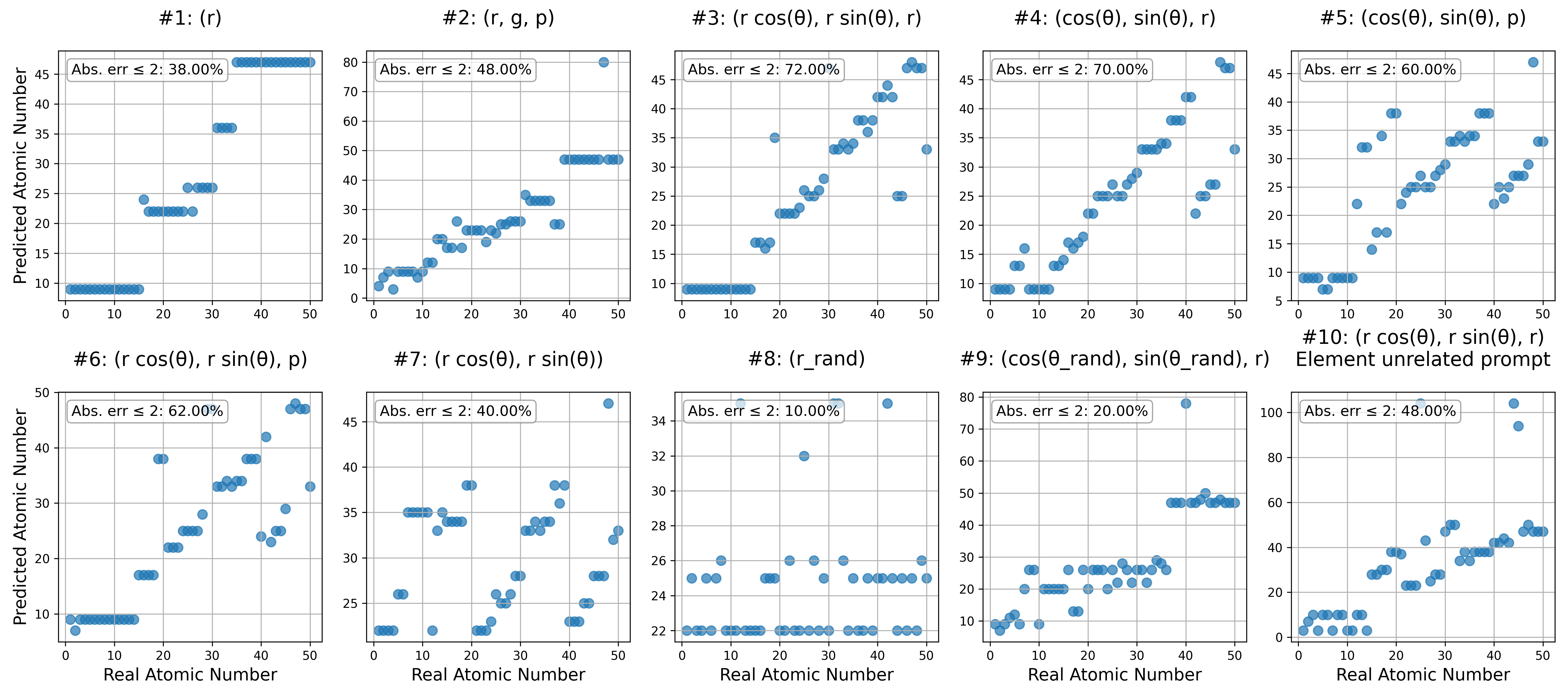}
    \caption{Residual Stream Patching Results for Layer 20 in Meta-Llama-3.1-70B. The model’s predictions are evaluated after replacing the residual stream of the `element' token at the last token position with the predicted residual stream \(\hat{\mathbf{h}}^{\text{pred}, (20)}_{0}\). }
    \label{fig:intervention}
\end{figure}

 Inspired by \citet{engels2024not}, we map the last-token residual streams $\mathbf{h}^{(l)} \in \mathbb{R}^k$ at layer $l$ to a geometric space $f(r, g, p)$, which encodes atomic number $r$, group $g$, and period $p$. To learn this mapping, we first reduce the dimensionality of the residual streams to 30 using PCA, denoted as $\mathbf{P}(\mathbf{h}^{(l)})$, and fit a linear projection using all elements except one held-out target:

$$
\mathbf{W}^{(l)}, \mathbf{b}^{(l)} = \arg \min_{\mathbf{W}', \mathbf{b}'} \sum_{i \ne 0} \left\| \mathbf{W}' \mathbf{P}(\mathbf{h}^{(l)}_{i}) + \mathbf{b}' - f_i \right\|_2^2
$$

where $\mathbf{W}^{(l)} \in \mathbb{R}^{d' \times 30}$, $\mathbf{b}^{(l)} \in \mathbb{R}^{d'}$, and $f_i = f(r_i, g_i, p_i)$ denotes the mapping of the \(i\)-th element in the geometric space.

To perform the intervention, we compute the centroid of the PCA-reduced residual streams for the remaining $N = K - 1$ elements:

$$
\bar{\mathbf{h}}^{(l)} = \frac{1}{N} \sum_{i \ne 0} \mathbf{P}(\mathbf{h}^{(l)}_i)
$$

then map it to the geometric space: $\mathbf{z} = \mathbf{W}^{(l)} \bar{\mathbf{h}}^{(l)} + \mathbf{b}^{(l)}$. Let $f_0 = f(r_0, g_0, p_0)$ denote the target element's embedding in the geometric space. The deviation $f_0 - \mathbf{z}$ is projected back to the residual stream space using the pseudo-inverse of $\mathbf{W}^{(l)}$, giving the predicted (intervened) residual stream:

$$
\hat{\mathbf{h}}^{\text{pred}, (l)}_0 = \mathbf{P}^{-1} \left( \bar{\mathbf{h}}^{(l)} + \left(\mathbf{W}^{(l)}\right)^+ (f_0 - \mathbf{z}) \right)
$$

\begin{wrapfigure}{r}{0.30\textwidth}
\centering
    \includegraphics[width=1\linewidth]{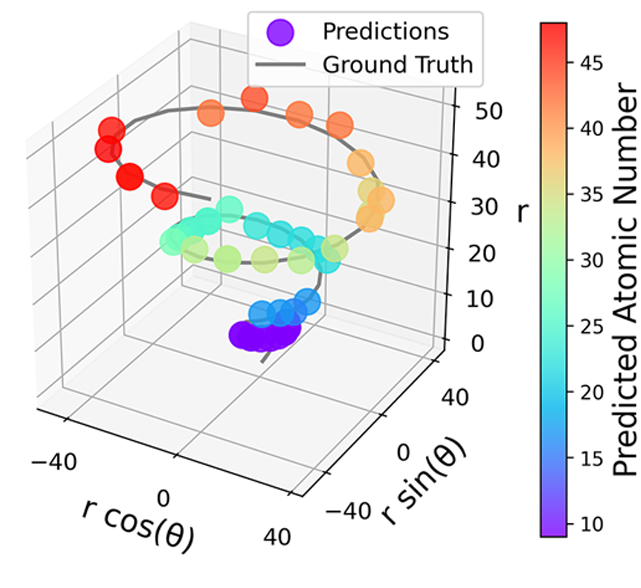}
    \caption{Predicted atomic numbers after intervention in 3D spiral space $(r \cos \theta, r \sin \theta, r)$. Colored points indicate the tokens with highest logits. 
}
    \label{fig:spiral}
\end{wrapfigure}

Importantly, the model never accesses the original residual stream of the target element; the predicted residual stream is computed solely from its geometric representation and the residual streams of other elements. During inference, we replace the residual stream of `element' (last token position) in the 20th layer\footnote{See Appendix \ref{app:layer} for details on intervention performance. Interventions become effective from layer 20 onward.} with $\hat{\mathbf{h}}^{\text{pred}, (20)}_{0}$, using the prompt \textit{`In the periodic table, the atomic number of element'}. We then evaluate whether the model can correctly output the target token without ever seeing its original residual stream.

We evaluate the effectiveness of different geometric spaces for interventions, including linear, 2D spiral, and 3D spiral geometries. Angular variables \(\theta = \frac{2\pi g}{18}\) are used to capture periodic relationships. To test the impact of disrupted geometry, two random spaces are introduced: in Space 8, atomic numbers \(r\) are shuffled; in Space 9, \(\theta\) is randomly permuted. Additionally, in Space 10, the prompt \textit{`In numbers, the Arabic numeral for number'} generates numbers 1–50, testing whether periodic patterns emerge without explicit element references. 

Effective residual stream patching suggests that the target space \(f(r, g, p)\): 1) retains sufficient information for accurate reconstruction during transformations with the residual stream space, and 2) preserves geometric structures similar to those in the residual stream space to ensure valid adjustments in the high-dimensional space.

Patching results for Meta-Llama-3.1-70B are shown in Fig. \ref{fig:intervention}, with detailed values in Table \ref{tab:lowdim_interventions} (Appendix). Results show that intervention can be applied in various geometric spaces, with some performing significantly better. Spaces such as \((\cos \theta, \sin \theta, r)\) and \((r \cos \theta, r \sin \theta, r)\) over \(70\%\) predictions of the atomic number have an absolute error within 2, suggesting the potential existence of latent 3D structures in LLMs resembling spirals or radial spirals. Fig. \ref{fig:spiral} illustrates the LLM's output post-intervention in 3D spiral geometry. Additional geometric analyses are in Appendix \ref{app:other_shape}. Random spaces and unrelated prompts perform poorly, reinforcing the correlation between embedding geometry and real-world knowledge structures. In a concurrent study, \citet{kantamneni2025language} observed spiral-like structures in number space with periods of 2, 5, 10, and 100, likely reflecting common human conventions in numerical representation. In contrast, our model exhibits a distinct 18-period spiral aligned with the periodic structure of chemical elements. This representation performs notably worse for ordinary numbers without elemental context (which aligns with their observation that the 18-period does not prominently emerge), indicating that such geometric patterns emerge from underlying physical or semantic regularities rather than arbitrary structures.


\begin{wrapfigure}{r}{0.35\textwidth}
\centering
    \includegraphics[width=1\linewidth]{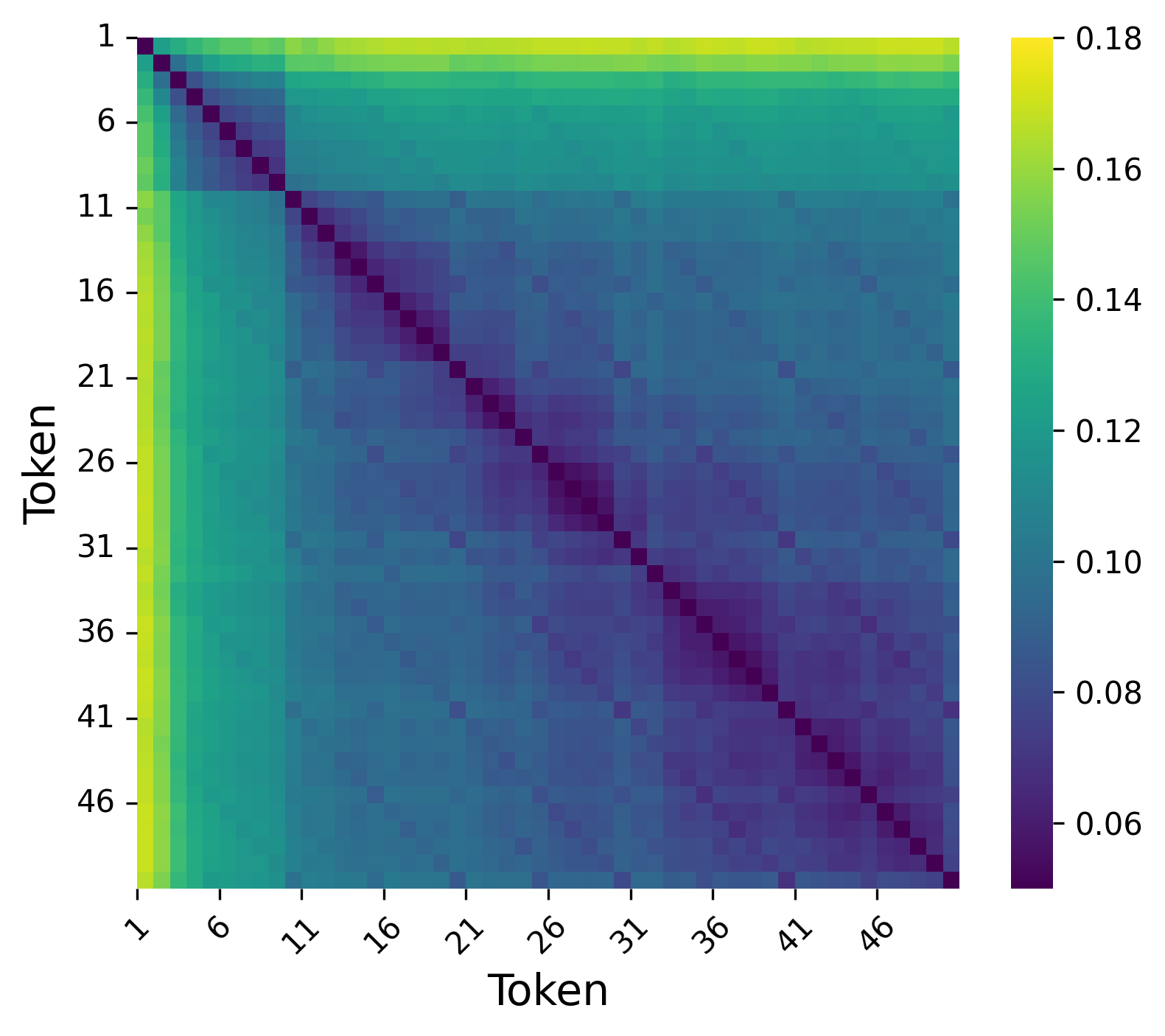}
    \caption{Euclidean distance heatmap of approximated vector representations for numeric tokens (1-50) in the hidden space of the last layer in Meta-Llama-3.1-70B.}
    \label{fig:distance}
\end{wrapfigure}

In the intervention experiments, it is actually not obvious whether a smaller numerical difference between the output token and the true value always implies smaller error. To investigate this, we project token IDs for numbers 1–50 into the last hidden layer using the pseudoinverse of the vocabulary projection matrix \( \mathbf{W}_{\text{vocab}}^+ \). This operation reconstructs an approximation of the hidden representations that would produce these token IDs as logits. Fig. \ref{fig:distance} shows that smaller numerical differences generally correspond to closer representations, while larger differences often result in inconsistent distances, reflecting the model's difficulty with numerical consistency over larger gaps. For instance, the vector for `1' is closer to `2' than to `5', while the distances between `10' and `40' is closer than between `10' and `21'. In the intervention, when the predicted value is close to the true value, hidden logits align well with true logits, suggesting higher accuracy. However, large numerical deviations cannot fully capture prediction errors, so we evaluate results using an absolute error threshold (\(\leq 2\)) in Fig. \ref{fig:intervention}, representing a small distance.


\section{Direct attribute recall}

In the previous section, we observed that elemental knowledge in LLMs forms a 3D spiral structure. Interestingly, although prompts mentioned only atomic numbers, the embeddings also reflected elemental groups, suggesting that LLMs retrieve both explicit and implicit attributes. To better understand these mechanisms, this section investigates direct attribute knowledge recall and Sec. \ref{sec:indirect_recall} will explores how LLMs access related but unprompted knowledge.

\subsection{From continuity to boundary sharpening}
\label{sub:probing}

Some elemental attributes, such as group and period, naturally exist in both categorical and numerical forms. This duality enables both classification and regression probing, allowing for direct comparisons that have been underexplored in prior work, which often focused exclusively on a single type.

To examine how LLMs access explicitly mentioned knowledge, we use the last-token residual stream from the continuation style prompt \(\mathbf{h}_j^{(l)} \in \mathbb{R}^k\) as the representation of attribute \(A_j\), and fit a linear probe to predict its corresponding values via:
\[
f_j^{(l)}(\mathbf{h}_j) = \mathbf{W}_j^{(l)} \mathbf{h}_j^{(l)} + \mathbf{b}_j^{(l)}
\]

For categorical attribute forms (\textit{e.g.}, Category, Group, Period), \(\mathbf{W}_j^{(l)} \in \mathbb{R}^{|C_j| \times k}\), \(\mathbf{b}_j^{(l)} \in \mathbb{R}^{|C_j|}\). Predictions are made by:
\[
\hat{y}^{(l)} = \arg\max_{c\in C_j} [f_j^{(l)}(\mathbf{h}_j^{(l)})]_c.
\]
For continuous attributes, we perform scalar regression by setting \(\mathbf{W}_j^{(l)} = \mathbf{w}_j^{(l)\top}\), \(\mathbf{w}_j^{(l)} \in \mathbb{R}^k\), \(\mathbf{b}_j^{(l)} \in \mathbb{R}\), yielding:
\[
\hat{y}^{(l)} = \mathbf{w}_j^{(l)\top} \mathbf{h}_j^{(l)} + b_j^{(l)}
\]
Probes are trained using 5-fold cross-validation on last-token residual streams. We use a linear Support Vector Machine (SVM) for categorical tasks and Support Vector Regression (SVR) with a linear kernel for continuous tasks. The resulting classification accuracies and regression \(R^2\) scores are shown in Fig.~\ref{fig:linear_probing}, with best-layer results provided in appendix \ref{app:best_layer}. 

Regression probes reveal that continuous numerical features are effectively represented in intermediate layers, as indicated by high $R^2$ values (while not reaching 1, see Appendix \ref{app:not_1}). These intermediate layers sometimes even outperform the final layers, suggesting that numerical knowledge is already encoded before the final output stage. This aligns with findings by \citep{meng2022locating}, which show that factual knowledge recall is already mediated by intermediate MLP layers.

\begin{figure}
\centering
    \includegraphics[width=1\linewidth]{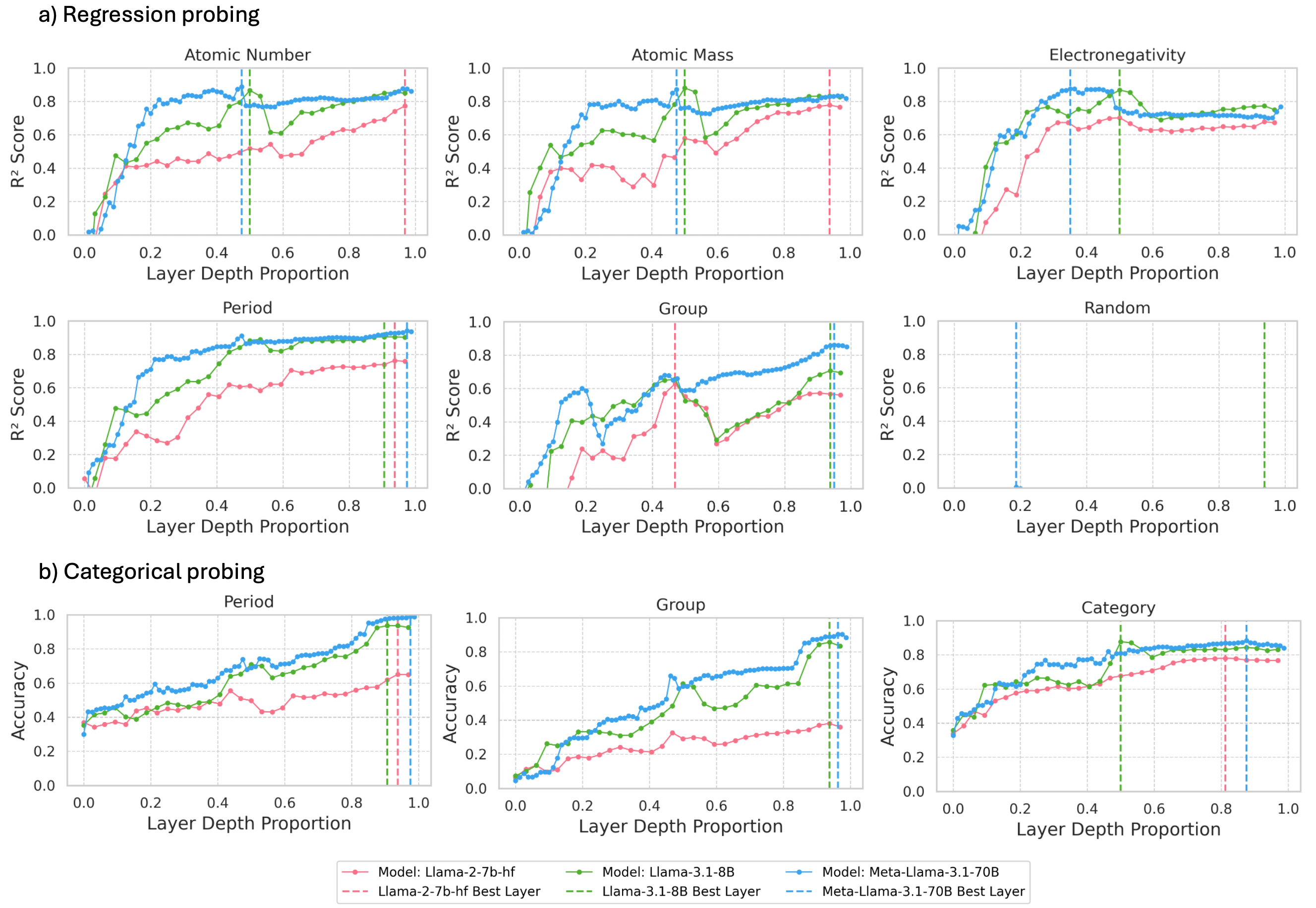}
    \caption{
    Linear probing results on last token across layers. (a) Regression (\(R^2\)) for numerical attributes and a random baseline. (b) Classification (accuracy) for categorical attributes. All results use 5-fold cross-validation on last-token residual streams.
    }
    \label{fig:linear_probing}
\end{figure}

In classification probes, intermediate layers perform similarly or even better than final layers for clearly distinct non-numerical categories (\textit{e.g.}, metal vs. non-metal), aligning with prior work \citep{nanda2023factfinding}. However, they significantly underperform in fine-grained numerical classification---\textit{e.g.}, Period accuracy drops from $\sim\!1.0$ (final) to $\sim\!0.7$, and Group from $\sim\!1.0$ to $\sim\!0.6$. 

This suggests that while intermediate layers already encode meaningful numerical structure, additional processing in later layers is required to sharpen boundaries and support accurate fine-grained classification. This aligns with intuition: later layers prepare for discrete token outputs, where clearer classification boundaries must emerge. As shown in Appendix Fig.~\ref{fig:confusion}, the confusion matrix from Layer 40 (70B middle layer) is not perfectly accurate, but most misclassifications fall near the diagonal—further demonstrating that intermediate layers encode coherent numerical structure, albeit with blurred categorical boundaries. These observations may provide useful insights for choosing between intermediate and later-layer embeddings in downstream tasks.

Notably, Llama2 7B shows low accuracy (<0.4) on Group classification compared to Llama3.1 8B (>0.8) (but similar performance in Group regression probing) potentially due to its single number tokenization (splitting numbers like `12' into `1' and `2'), which may cause confusion between the representations of output tokens like `12' and `1'. In contrast, Llama 3 uses separate tokens for numbers below 1000.


 \begin{wrapfigure}{r}{0.4\textwidth}
\centering
    \includegraphics[width=1\linewidth]{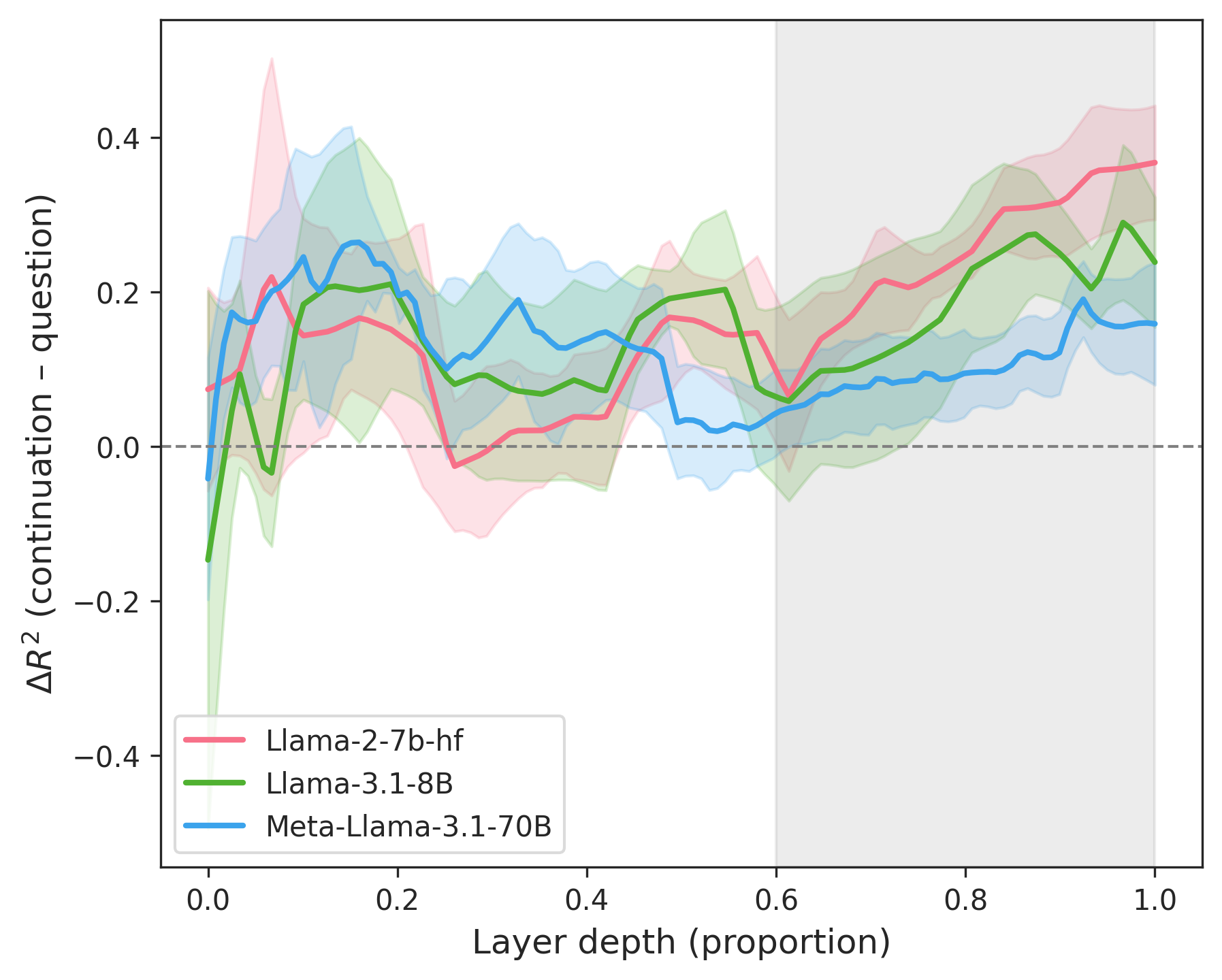}
    \caption{Average \(\Delta R^2\) across five attributes, with with 95\% confidence interval shaded. \(\Delta R^2 = R^2_{\text{cont}} - R^2_{\text{ques}}\).}
    \label{fig:question_vs}
\end{wrapfigure}

\subsection{Higher Language Sensitivity in Later Layers}
\label{sub:language}

The sharpening of numerical representations into categorical boundaries in later layers suggests that these layers might be shaped by the expected output tokens. This raises a question: does the linguistic structure influence the factual representations across layers?

We compared question-style and continuation-style prompts using linear regression probes. Continuation prompts generally lead to direct generation of fact-related tokens, whereas question-style prompts tend to introduce syntactic fillers (\textit{e.g.}, `The') and are more influenced by surface language patterns.

Fig.\ref{fig:question_vs} reports the average delta $R^2$ across five attributes, with per-attribute results shown in Fig.~\ref{fig:question_vs_con} (Appendix). As analyses in earlier sections show stronger semantic signals and higher $R^2$ in mid-to-late layers, we focus on depths 0.6–1.0. $\Delta{R^2}$ increases in the mid-to-late layers, indicating a growing gap between prompt types. Among the 15 attribute–model combinations (3 models × 5 attributes), 12 show a significant increasing trend (FDR-corrected $p < 0.05$), with a median Mann–Kendall $\tau$ of 0.55 (Appendix \ref{app: question_vs}). 

The results indicate that, as depth increases, question prompts become progressively less effective than continuation prompts at encoding factual attributes, hinting that the prompt’s linguistic structure exerts a stronger influence on representations in deeper layers. Interestingly, larger models show a slower increase in $\Delta R^2$ across layers than smaller models, suggesting they maintain more stable factual representations across prompt types and thus exhibit a smaller distinction between continuation and question prompts.

The rising $\Delta R^2$ suggests that deeper layers increasingly blend factual content with linguistic structure to prepare the final tokens. To further test this, we applied the logit lens \citep{nostalgebraist2020logitlens} and tuned-lens \citep{belrose2023eliciting}. These analyses estimate the token distribution each layer would produce if decoding were halted at that depth, and show that the correct numerical token becomes highly ranked only in the later layers (Appendix~\ref{app:len}). Complementary attention statistics (Appendix~\ref{app: attention}) reveal that mid-layers focus tightly on the factual token, whereas later layers spread attention over a wider context—patterns consistent with increased syntactic and contextual integration.


\section{Indirect attribute recall}
\label{sec:indirect_recall}
In the previous section, we analyzed direct recall of explicitly mentioned attributes across layers. Our earlier geometric analysis showed that LLMs can also recall related attributes that are not explicitly mentioned. In this section, we explore how related but unmentioned attributes are recalled.

\subsection{Middle Layers excel at indirect recall}
\label{sub: middel_in}

\begin{figure}
\centering
    \includegraphics[width=1\linewidth]{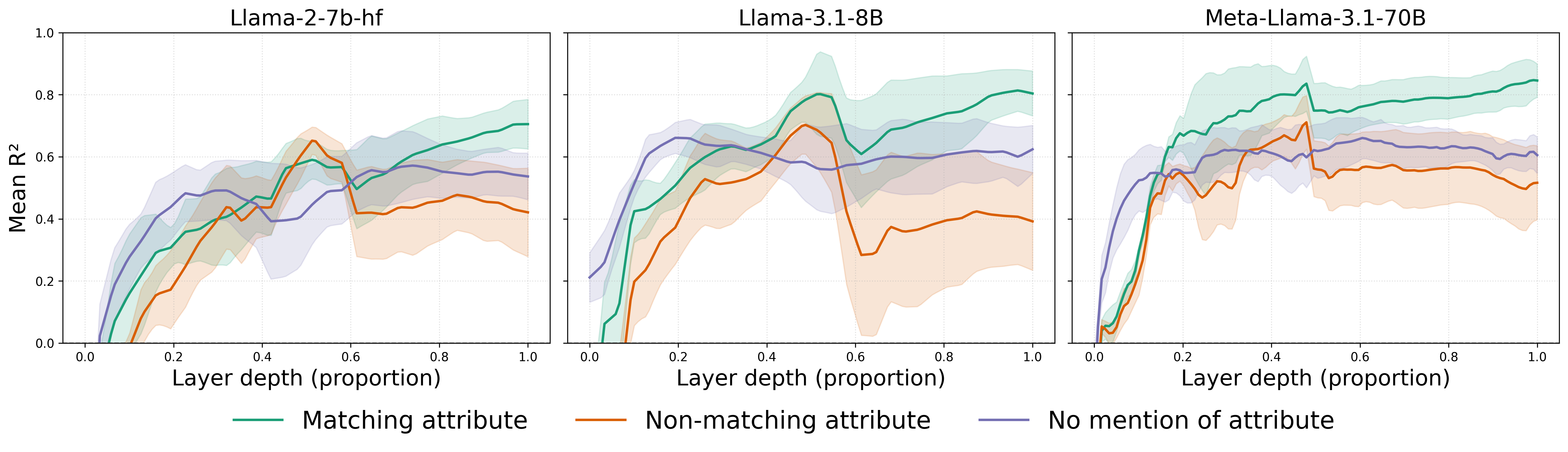}
    \caption{Average $R^2$ scores from regression probing across layers for three prompt types: matching, non-matching, and no-mention. All probes predict a fixed target attribute $A_{j_1}$; no-mention uses element token residual streams before any attribute appears. Shaded areas show 95\% confidence intervals.
}
    \label{fig:non_matching_3}
\end{figure}

We conducted experiments using linear probing to examine the relationships between distinct attributes. Specifically, we extracted last-token residual streams from continuation prompts that mention attribute $A_{j_1}$ (matching) or a different attribute $A_{j_2}$ (non-matching). We also extracted the residual stream at the element token position, before any attribute is introduced (no mention). Separate probes were trained for each residual stream dataset, always using labels of attribute $A_{j_1}$ as targets. To avoid confounding factors, we selected six attribute pairs without direct linear relationships for non matching probe (see Appendix \ref{par:pair_choosing}).  Average $R^2$ curves for all attributes are shown in Fig.~\ref{fig:non_matching_3}; detailed case-wise linear probing results appear in Appendix Figs.~\ref{fig:regression_ele} and \ref{fig:matching_detail}.

Attribute information was detectable across all prompt styles. Intuitively, matching prompts should perform best by providing explicit cues, no-mention comes next as it relies on inference, and non-matching prompts perform worst due to misleading signals. Surprisingly, at intermediate layers (around 0.5 depth), non-matching prompts yielded higher linear $R^2$ scores than no-mention prompts, suggesting stronger inter-attribute interactions at these depths. This may reflect entangled representations between related attributes, which we analyze further in Sec.~\ref{sec: relationship}.

Beyond 60\% depth, performance follows the expected trend: matching > no-mention > non-matching. The gap between matching and non-matching prompts increases steadily from 0.6 to 1.0 depth. Across 15 model–attribute tests, 14 exhibited statistically significant divergence (FDR corrected $p<0.05$), with a median Mann–Kendall τ of 0.77 (Appendix Fig.\ref{fig:delta_nonmatching}, Table \ref{tab:mk_results}). It suggests that attribute representations become more specialized and context-sensitive in deeper layers. Further analyses in Sec. \ref{sec: relationship} provide a more direct explanation, examining how structural relationships between attributes contribute to this layered specialization.

The fact that the `no-mention' prompts perform best in the early layers may seem counterintuitive; however, this is likely because, unlike the other two scenarios, in the `no-mention' case, the last token is the element itself, which may aid recall. In contrast, matching prompts extract residual streams at the final token (such as `is'), requiring holistic semantic understanding. As layer depth increases, semantic clarity improves, enhancing explicitly mentioned attributes and reversing this initial trend.


\begin{wrapfigure}{r}{0.4\textwidth}
\centering
    \includegraphics[width=1\linewidth]{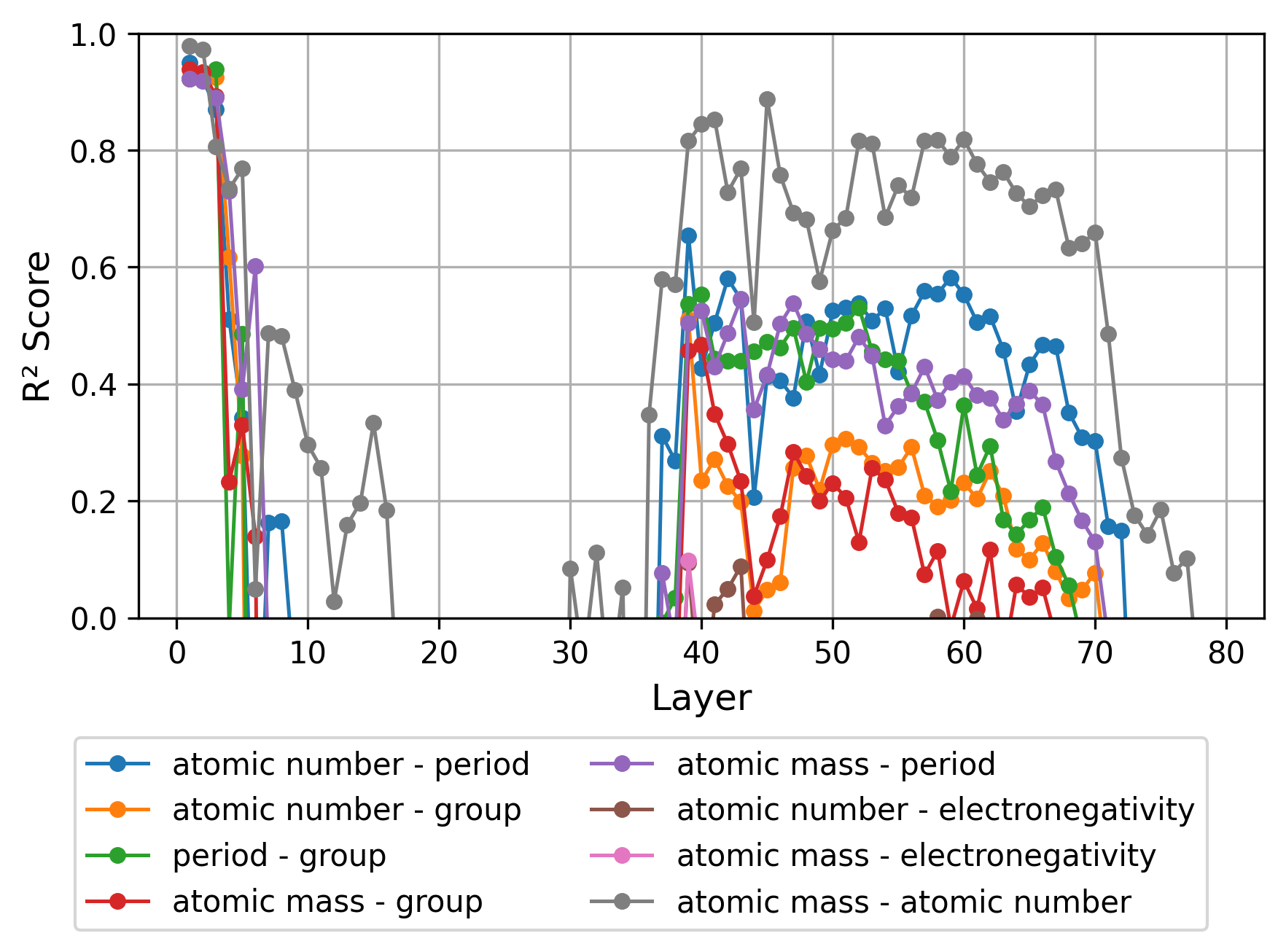}
    \caption{$R_2$ scores across layers in Meta-Llama-3.1-70B for linear mappings between attribute pairs using the final residual stream from a fixed prompt.}
    \label{fig:relationship}
\end{wrapfigure}

\subsection{Stronger linear correlations in middle layers}
\label{sec: relationship}

To explicitly capture relationships between attribute representation, we train a linear mapping from the representation of attribute $A_{j1}$ to attribute $A_{j2}$ at each model layer. Specifically, we utilize the final residual streams from a fixed prompt template (after applying PCA to reduce the dimensionality to 20). The mapping performance is evaluated using $R^2$ scores obtained via 5-fold cross-validation.

Fig. \ref{fig:relationship} illustrates the variation of $R^2$ scores across layers for different attribute pairs. In early layers, $R^2$ scores are high; however, this observation alone does not necessarily indicate meaningful attribute-level relationships, as initial representations are predominantly sensitive to token-level similarity (see t-SNE analysis in Fig \ref{fig:vis}). Due to the use of a fixed input template, the resulting inputs exhibit substantial token-level overlap.

In the intermediate layers, where concept-level understanding is evident (as shown by t-SNE and linear probing), we observe a peak in $R^2$ scores. This indicates that even simple linear models can effectively capture relationships between different attributes, reflecting their connection in the learned representation space. This also explains why prompts with non-matching attributes outperform those with no attribute mention at these layers in the last section Sec \ref{sub: middel_in}. In deeper layers, $R^2$ scores decline, suggesting a shift toward specialized representations. Similar conclusions from the linear probing weight analysis further support this, as shown in Appendix \ref{app:probing_weight}.

\section{Related work}
\label{sec:related work} 


\textbf{Linear representation hypothesis.} The linear representation hypothesis suggests that neural networks encode high-level features as linear directions in activation space, enabling easier interpretation and manipulation \citep{nanda2023b}. Probing, introduced by \citep{alain2016understanding}, assesses feature encoding in models and builds on findings in word embeddings like GloVe and Word2Vec, which capture semantic relationships through linear structures \citep{mikolov2013, pennington2014}. Empirical support spans various contexts, including spatial and temporal representations \citep{gurnee2023language}, sentiment analysis \citep{tigges2023linear}, task-specific features \citep{hendel2023context}, and broader relational structures \citep{hernandez2023linearity}.

\textbf{Non-linear Representations.} Although the linear representation hypothesis offers insights into neural network representations, studies have highlighted its limitations and emphasized the significance of non-linear structures. Non-linear structures, such as the `pizza' and `clock' patterns \citep{nanda2023progress, zhong2024clock}, and circular representations observed in tasks like predicting days or months using modular arithmetic prompts \citep{engels2024not}, reveal the complexity of these representations. Following this, we identify a 3D spiral in element representations aligned with periodic trends, suggesting geometric organization of knowledge in LLMs.

\textbf{Intermediate layers matter.}
Recent studies underscore the importance of intermediate layers in LLMs, emphasizing their role in producing more informative representations for downstream tasks compared to final layers \citep{skean2025layer, kavehzadeh2024sorted, ju2024large, liu2024fantastic}. These layers are crucial for encoding abstract knowledge, enabling advanced capabilities like in-context learning and transfer learning, which are vital for understanding and optimizing LLMs \citep{zhang-etal-2024-investigating}. Additionally, intermediate layers exhibit distinct patterns of information compression and abstraction, such as reduced entropy, allowing them to efficiently represent complex inputs \citep{doimo2024representation, yin2024entropy}. Building on these findings, our probe results show that intermediate layers encode knowledge in a continuous form, while sharper categorical boundaries emerge in later layers.

\textbf{Factual recall.}
\citet{meng2022locating} showed that early MLP layers at the entity token are key to recalling factual associations, while later attention layers propagate this information to the output. \citet{geva2023dissecting} expanded this into a three-stage process—enrichment, transfer, and extraction, revealing subject tokens carry multiple implicit attributes. \citet{nanda2023factfinding} further validated this through detailed circuit analysis, demonstrating entity token representations linearly encode categorical attributes. Building on this, we investigate how multiple related attributes are jointly represented, and show that intermediate layers encode multiple related attributes in structured and overlapping subspaces, forming geometry-aware representations.

\section{Discussion and conclusions}

This study reveals LLMs, despite being trained solely on textual data, develop internal representations that reflect the structured geometry of scientific knowledge. We uncover a 3D spiral structure in LLMs that organizes chemical elements in alignment with the structure of periodic table, suggesting that the model implicitly captures domain-specific regularities without explicit supervision. Through probing experiments, we show that middle layers encode continuous, overlapping attribute subspaces suitable for coarse categorization, while deeper layers sharpen decision boundaries and integrate linguistic structure. Moreover, we find that related attributes are strongly linearly associated in middle layers, enabling indirect recall. 

These findings demonstrate that LLMs represent symbolic knowledge not as isolated facts but as geometry-aware, interwoven manifolds. We hope this work inspires further investigation into how LLMs represent and reason about scientific knowledge, such as materials property prediction, and informs the design of downstream embedding-based tasks. We believe interpretability in LLMs is essential for AI safety, reducing unintended behaviors and building trust. Understanding how knowledge is stored and recalled across layers can inspire more interpretable, efficient models, advance knowledge editing and scientific discovery.

\textbf{Limitations.}
Our prompts are focused on chemical elements, while ideal for their structured attributes, may not extend to domains with more abstract features. The hypothesis-driven validation of geometric structures may oversimplify LLMs' non-linear interactions.

\section*{Code Availability}
The code required to reproduce the results presented in this paper is available at \url{https://github.com/tldr-group/LLM-knowledge-representation} with an MIT license agreement. 

\section*{Acknowledgements}
This work was supported by the Imperial Lee Family Scholarship awarded to G.L. We would like to thank the members of the TLDR group for their valuable comments and insightful discussions.



\bibliographystyle{plainnat}  
\bibliography{references}


\newpage

\input{appendix}



\end{document}

%% file: appendix.tex
\appendix
\renewcommand{\thefigure}{\thesection.\arabic{figure}}
\renewcommand{\thetable}{\thesection.\arabic{table}}

\makeatletter
\@addtoreset{figure}{section}
\@addtoreset{table}{section}
\makeatother

\section{Last token residual stream collection}
\label{app:activation_collection}

For each layer \(l\), we collect last-token residual streams \(\mathbf{h}_{i,j,k}^{(l)}\) from prompts \(p_{i,j,k}\) across all elements and templates:
\[
\mathbf{h}_{i,j,k}^{(l)} = f^{(l)}\bigl(p_{i,j,k}\bigr) \in \mathbb{R}^{T \times d},
\]
where \(f^{(l)}\) denotes the layer-\(l\) transformation, \(T\) is the token length, and \(d\) is the hidden dimension. The initial residual stream \(\mathbf{h}_{i,j,k}^{(0)}\) is obtained by embedding the prompt through an embedding layer \(E_0\), followed by processing through \(L\) Transformer layers. Each layer applies multi-head attention and a feedforward network with residual connections and layer normalization:  
\[
\mathbf{h}{\prime}^{(l)}{i,j,k}
= \mathbf{h}^{(l-1)}{i,j,k}
\;+\; \mathrm{Attention}\!\bigl(\mathrm{LayerNorm}(\mathbf{h}^{(l-1)}_{i,j,k})\bigr)
\]
\[
\mathbf{h}^{(l)}{i,j,k}
= \mathbf{h}{\prime}^{(l)}{i,j,k}
\;+\; \mathrm{FFN}\!\bigl(\mathrm{LayerNorm}(\mathbf{h}{\prime}^{(l)}_{i,j,k})\bigr)
\]

Here, \( \mathbf{Q} \), \( \mathbf{K} \), and \( \mathbf{V} \) represent the query, key, and value matrices used in multi-head attention to compute token-to-token interactions. Finally, \(\mathbf{h}_{i,j,k}^{(L)}\) is mapped to the vocabulary space using the vocabulary head \(\mathbf{W}_{\text{vocab}}\) to produce logits:  
\[
\text{logits}_{i,j,k} = \mathbf{h}_{i,j,k}^{(L)} \mathbf{W}_{\text{vocab}}
\]  

By analyzing last-token residual streams \(\mathbf{h}_{i,j,k}^{(l)}\) across layers, we investigate how attributes are represented in the model's hidden states.

\section{Intervention outcomes in geometric recall}
\label{recall}

\subsection{Layer-wise performance evaluation}
\label{app:layer}

\begin{figure*}[h]
\centering
    \includegraphics[width=0.7\linewidth]{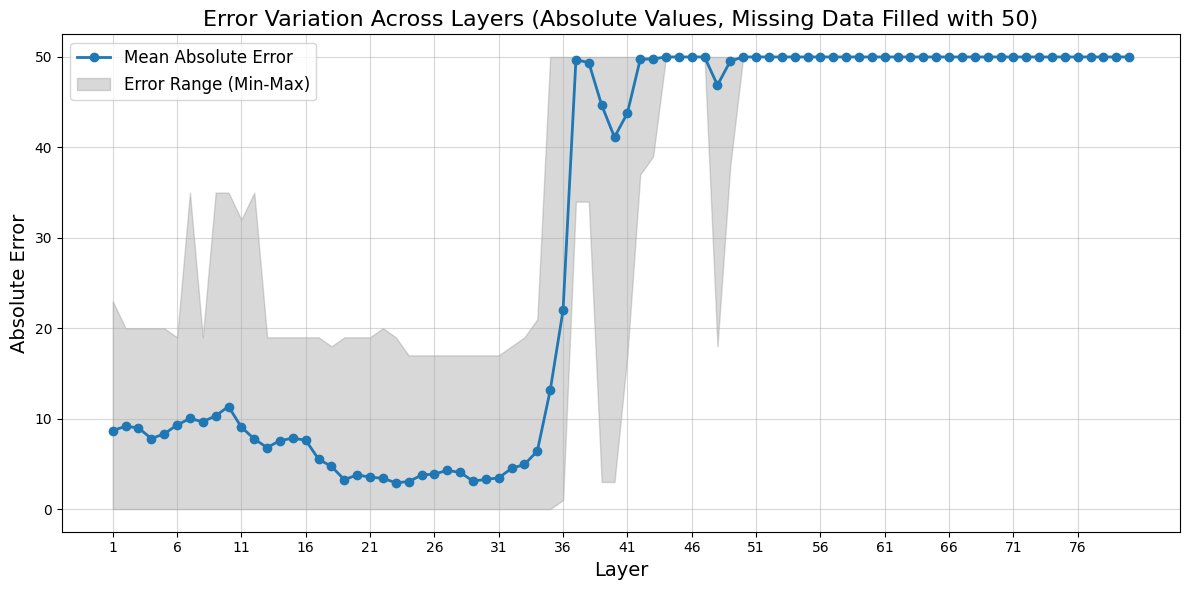}
     \caption{Variation of Absolute Prediction Errors Across Layers with Intervention. The plot shows the mean absolute error (MAE) for each layer, along with the minimum and maximum error range represented by the shaded region. Missing data points were replaced with a value of 50 before computing the absolute errors. }

    \label{fig:inter_error}
\end{figure*}

Fig. \ref{fig:inter_error} illustrates the prediction error across layers when the residual stream of the last token across layers is replaced with the predicted residual stream derived from the geometric space \(f(r, g, p) = (r\cos\theta, r\sin\theta, r)\). In the early layers, errors gradually decrease because the model has not yet captured semantic information, and the geometric space is still being constructed. The continuous decline in error reflects the model’s growing ability to capture semantic information and progressively build a coherent geometric representation. By layer 20, the error stabilizes, indicating that these layers effectively encode the periodic and geometric relationships between atomic properties such as atomic number, group, and period.

However, beyond layer 30, the error increases sharply as the model begins outputting non-numeric tokens (replaced with an error value of 50 in the plot, corresponding to the maximum possible error given the 50 elements and atomic numbers). This can be attributed to two factors. First, if the numeric token is not the first output, generating the correct answer requires residual streams across all token positions. But only the last residual stream was replaced. Intervening too late disrupts the established flow of residual streams at other positions, which have already determined the output content. Second, it is also likely due to the model shifting its focus from geometric relationships to higher-level abstractions or context-dependent reasoning in the later layers. Therefore, for the intervention experiments on geometric relationships, we selected layer 20 as it balances effective encoding of geometric relationships and minimizes disruption to the model's output process.

\subsection{Detailed evaluation of geometric spaces}
\label{app:other_shape}

The primary evaluation criterion used in the main text is the absolute error threshold (\(\leq 2\)), as discussed in detail in Sec.\ref{sec:geometric}. This metric was chosen because it better captures the accuracy of residual stream interventions. However, other metrics, such as \(R^2\), Pearson correlation, and qualitative mapping fidelity, also provide valuable insights. These complementary results are summarized in Table \ref{tab:lowdim_interventions}.

\begin{table*}[h!]
\centering
\footnotesize
\resizebox{\textwidth}{!}{
\begin{tabular}{@{}c p{3cm} cccc c@{}}
\toprule
\textbf{\#} & \textbf{Space} & \textbf{Description} & \shortstack{\textbf{R\textsuperscript{2}}} & \shortstack{\textbf{Pearson} \\ \textbf{Correlation}} & \shortstack{\textbf{Percentage of} \\ \textbf{Abs. err \(\leq 2\) }} & \textbf{Mapping Fidelity} \\ 
\midrule
1 & \(r\) & Linear structure along atomic number. & 0.8863 & 0.9591 & 38.00\% & Moderate \\ 
2 & \((r, g, p)\) & 3D cartesian grid. & 0.8060 & 0.9191 & 48.00\% & Moderate \\
3 & \((r \cos \theta, r \sin \theta, r)\) & 3D radial spiral structure. & 0.8162 & 0.9035 & 72.00\% & High\\
4 & \((\cos \theta, \sin \theta, r)\) & 3D spiral structure. & 0.7596 & 0.8813 & 70.00\% & High \\
5 & \((\cos \theta, \sin \theta, p)\) & 3D periodic wave-like structure. & 0.5106 & 0.7174 & 60.00\% & Moderate \\
6 & \((r \cos \theta, r \sin \theta, p)\) & 3D periodic lattice with radial dependencies. & 0.6719 & 0.8240 & 62.00\% & Moderate \\
7 & \((r \cos \theta, r \sin \theta)\) & 2D radial structure. & -0.1391 & 0.1481 & 40.00\% & Low \\
8 & \(r_\text{random}\) & Random linear structure. & 0.0075 & 0.1503 & 10.00\% & Low \\
9 & \((\cos(\theta_\text{random}), \sin(\theta_\text{random}), r)\) & Randomized spiral. & 0.6358 & 0.8465 & 20.00\% & Low \\
10 & \((r \cos \theta, r \sin \theta, r)\) & Element unrelated prompts & -0.4910 & 0.7215 & 48.00\% & Low \\
\bottomrule
\end{tabular}}
\caption{Performance of different low-dimensional spaces for residual stream intervention. Each space represents a unique pattern, with results assessed using \(R^2\), Pearson correlation, and percentage of predictions within absolute error \(\leq 2\).}
\label{tab:lowdim_interventions}
\end{table*}

\begin{figure*}[h]
\centering
    \includegraphics[width=0.7\linewidth]{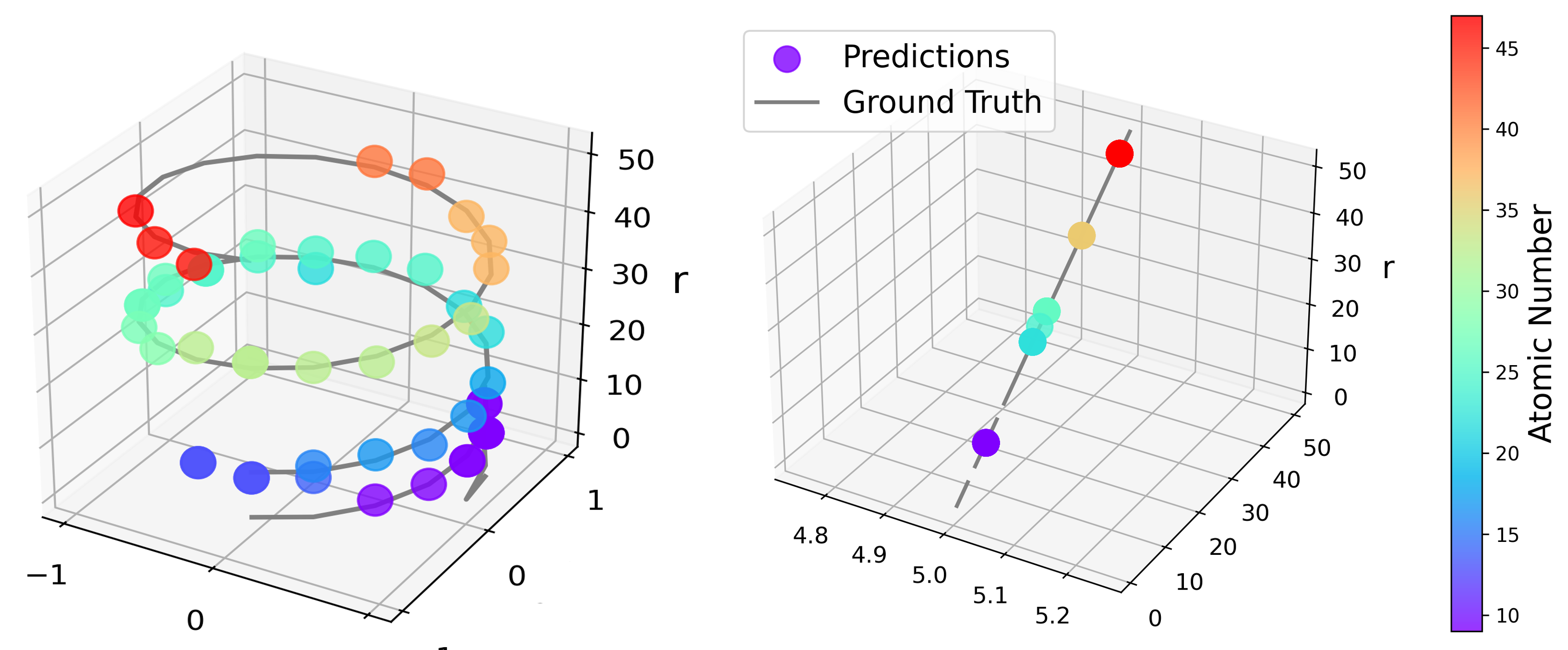}
     \caption{Helix and Linear Structure in the Geometric Space from Intervention Experiments. The figure shows predictions (colored points) and their alignment with the ground truth (gray line).}

    \label{fig:inter_linear}
\end{figure*}

In the main paper, we demonstrate two geometric space intervention results; however, other shapes can also be extracted. Fig. \ref{fig:inter_linear} shows the extracted linear structure from interventions. While the alignment of points along a straight path indicates the presence of a linear structure, the overlapping points suggest its limitations in distinguishing atomic number. Compared to more expressive shapes like spirals, linear structures may struggle to effectively capture periodic or distinct features.

\section{Attention map detailed results}
\label{app: attention}

\begin{figure}[h]
\centering
    \includegraphics[width=1\linewidth]{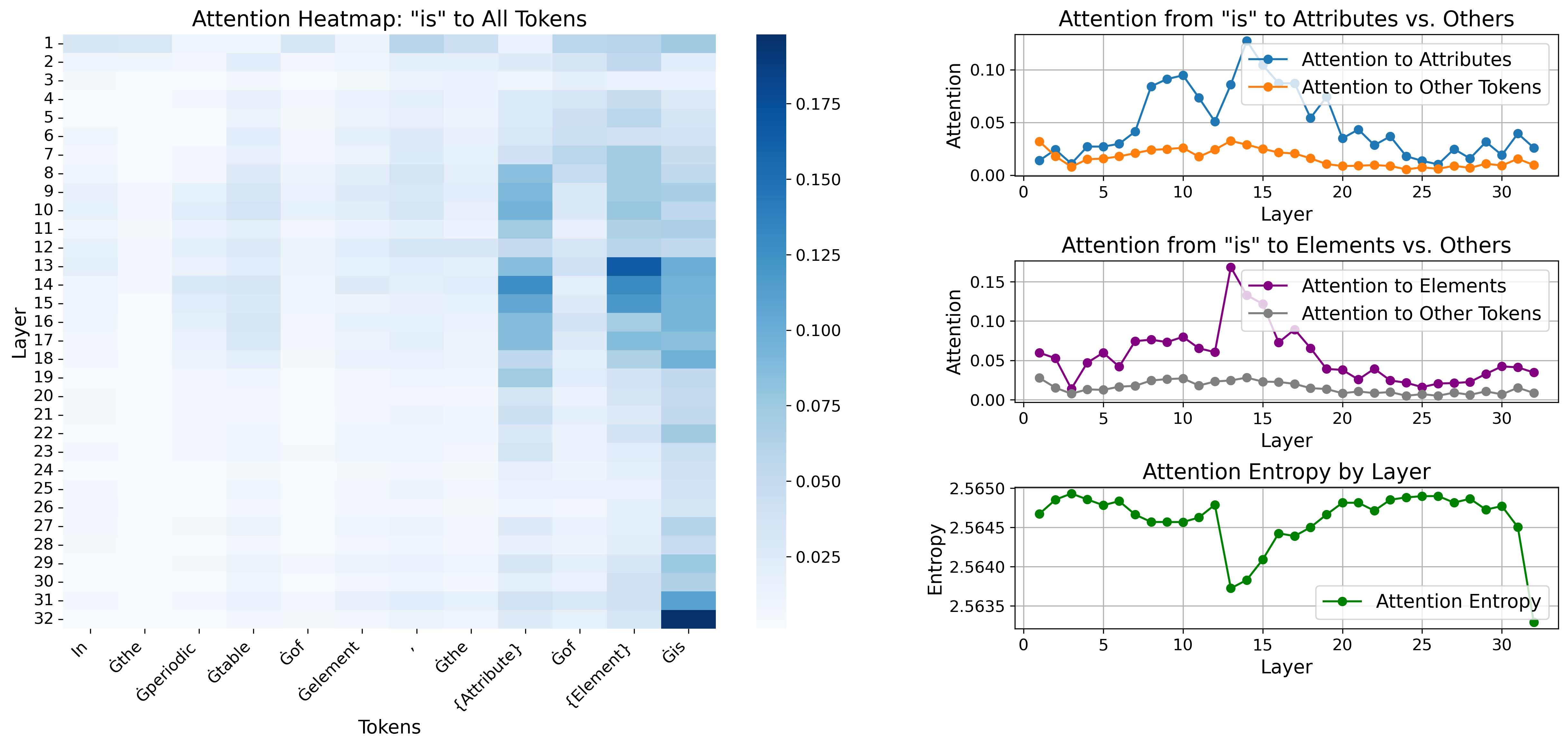}
     \caption{Average attention distribution analysis of the 32-layer Meta-Llama-3-8B model across transformer layers, based on prompts, `In the periodic table of elements, the  \( A_j \)  of \( X_i \) is,' where \( A_j \) is an attribute (period or group) and \( X_i \) is an element. The heatmap (left) shows average attention from `is' to all tokens, while line plots (right) depict attention to target tokens (e.g., element and attribute), average attention to other tokens, and attention entropy. Intermediate layers focus on significant token relationships with higher entropy, while later layers (excluding the final layer) show evenly distributed attention and lower entropy, reflecting a shift to broader context integration.}

    \label{fig:attention}
\end{figure}

To investigate how the model prioritizes different parts of the input text, we conducted a preliminary analysis using the 32-layer Meta-Llama-3-8B model. We adopted the attribute \( A_j \), Period and Group, and iterated over \( X_i \), consisting of 50 elements, using the prompt template: `In the periodic table of elements, the \( A_j \) of \( X_i \) is.' These prompts were input into the language model, and we analyzed the average attention across all attention heads in each transformer layer from the token `is' to all other tokens. The averaged results across different prompts are presented in Fig.\ref{fig:attention}.

The results indicate that in the intermediate layers, where entropy is relatively high, there is a noticeable concentration of attention from the token `is' to attribute and element tokens. This suggests that these intermediate layers focus more on tokens within the sequence that have a significant impact on the output. In contrast, the later layers, which exhibit lower entropy (with the exception of the final layer), show a more evenly distributed attention pattern. This pattern implies that the model transitions from focusing on specific token relationships to integrating broader context, thereby finalizing its interpretation for a cohesive output.

\section{Logit len and tuned len}
\label{app:len}
\begin{figure}[h]
\centering
    \includegraphics[width=0.5\linewidth]{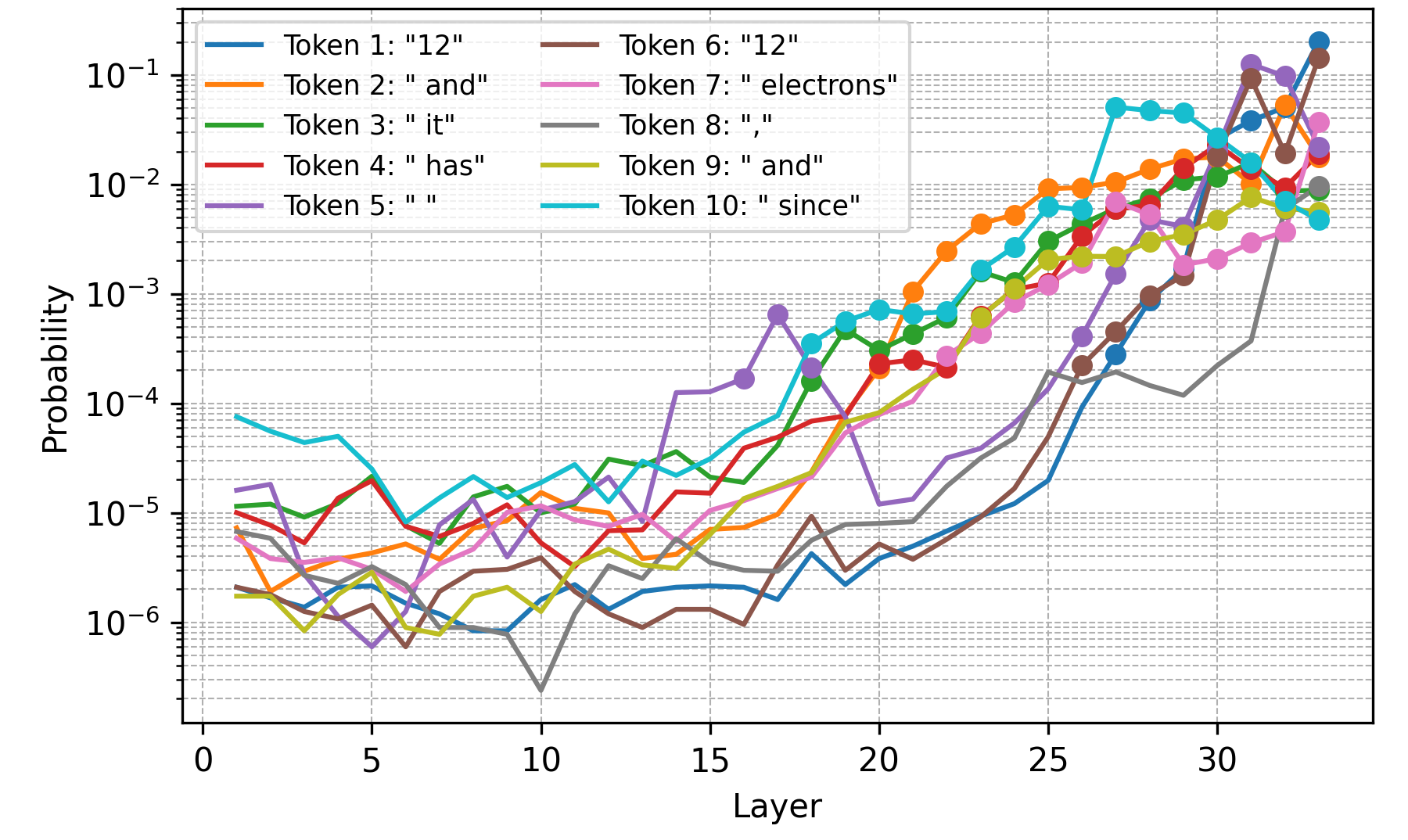}
    \caption{Probability of the target token $t_{\text{target}}$ across layers in Meta-Llama-3-8B for the prompt “The atomic number of Mg is”. Following the logit lens framework, each line shows $t_{\text{target}}$’s probability derived from intermediate layer logits. Probabilities are computed by iteratively re-running the model with the next token added. Markers indicate layers where $t_{\text{target}}$ ranks in the top 50 most probable tokens.}
    \label{fig:token_prob}
\end{figure}

We input the prompt `The atomic number of Mg is ' and analyzed the token probabilities at each layer using logit-len. By normalizing the final token’s hidden state with LayerNorm and applying the vocabulary head followed by softmax, we obtained the top-ranked tokens directly output by each layer. In each layer, we extracted the probability of the target token, \( t_{\text{target}} \) — the output token from the last layer, and checked if it ranked within the top 50 most probable tokens for that layer. The results are shown in Fig. \ref{fig:token_prob}.

In the early layers, the probability of the target token has not shown an upward trend, indicating these layers neither strongly predict the target tokens nor significantly refine their probabilities. In contrast, probabilities gradually increase in the later layers, highlighting their role in refining and finalizing predictions. Although crucially, there don’t appear to be any hard boundaries between these distinct activities and the model smoothly transition from one to the next. The markers, concentrated in later layers, suggest that while intermediate layers store factual knowledge, they are not yet attempting to articulate it in language form.

Notably, the distribution of `Top 50' markers varies by token type. Tokens with lower contextual complexity, such as spaces, `and,' or `since,' have their markers in earlier layers. In contrast, knowledge-based tokens, like `12,' require deeper processing and appear in much later layers. This suggests that while intermediate layers encode factual concepts, they are likely focused on tasks other than linguistic articulation, which primarily develops in the later layers.

\begin{figure}[h]
\centering
    \includegraphics[width=0.5\linewidth]{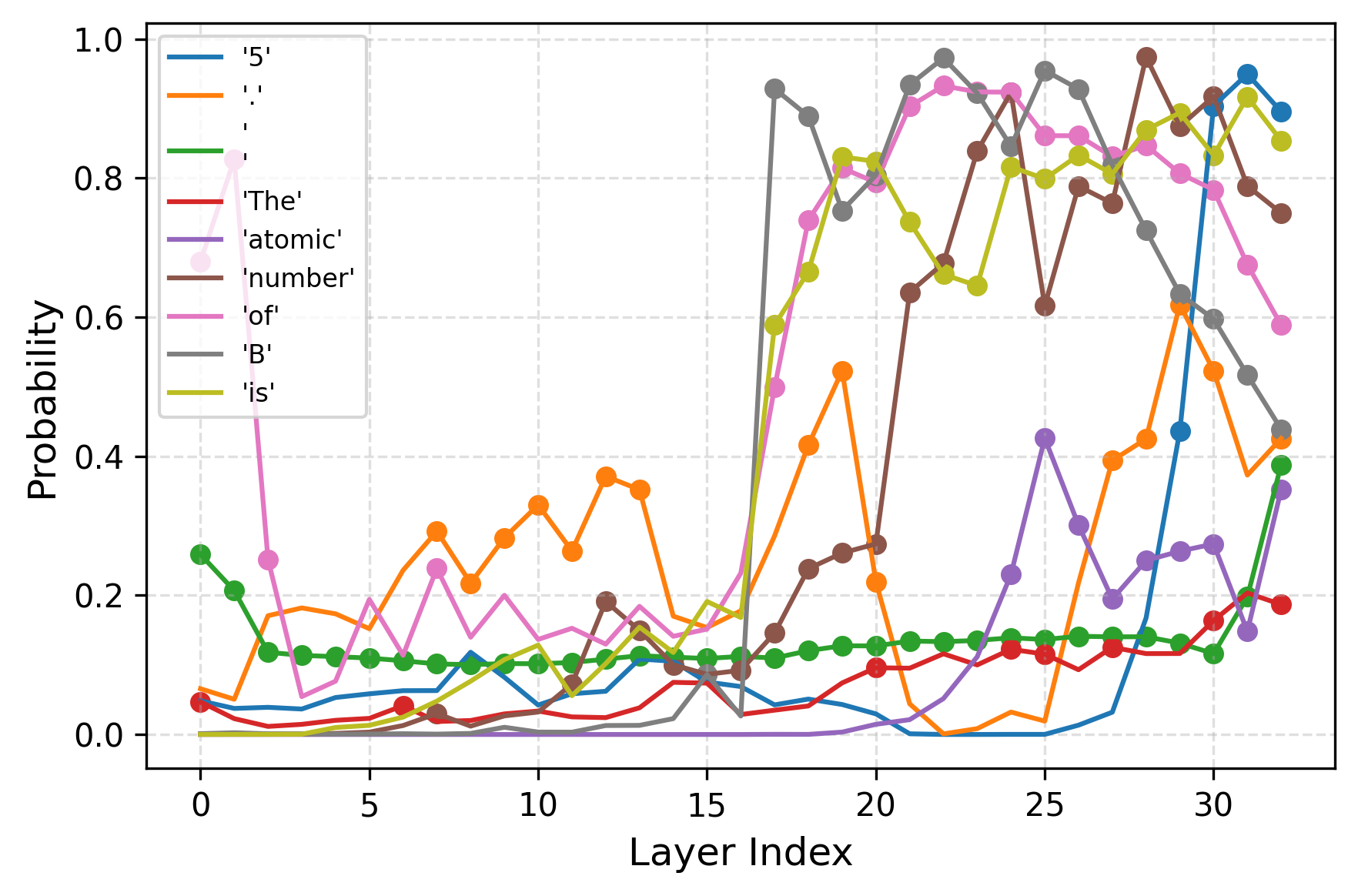}
    \caption{Probability of the target token $t_{\text{target}}$ across layers in Meta-Llama-3-8B for the prompt “The atomic number of Mg is”. Each line shows $t_{\text{target}}$’s probability computed using a tuned lens—a learned linear probe trained to decode intermediate hidden states. Probabilities are obtained by iteratively re-running the model with the next token appended. Markers indicate layers where $t_{\text{target}}$ ranks among the top prediction.}
    \label{fig:tuned_len}
\end{figure}

We also performed additional experiments using tuned lens in Fig \ref{fig:tuned_len}. The results consistently show a similar trend: tokens such as “of” and “is” achieve high prediction accuracy at intermediate or earlier layers, while more factual or critical token (e.g., blue line “5” only becomes the top prediction at layer 29)—require deeper processing. 

\section{Attribute representations overlap in intermediate layers}

\subsection{Probing weights analysis}
\label{app:probing_weight}
As outlined in Sec. \ref{sub:probing}, we trained a linear model for each attribute \( A_j \) at each layer \( l \), yielding a weight vector \( \mathbf{w}_j^{(l)} \) that represents how attribute \( A_j \) is stored in the residual stream space of layer \( l \). To analyze attribute relationships across layers, we computed the cosine similarity between weight vectors of different attributes using continuation-style residual stream sets to minimize language pattern influence.

\begin{figure}
\centering
    \includegraphics[width=0.6\linewidth]{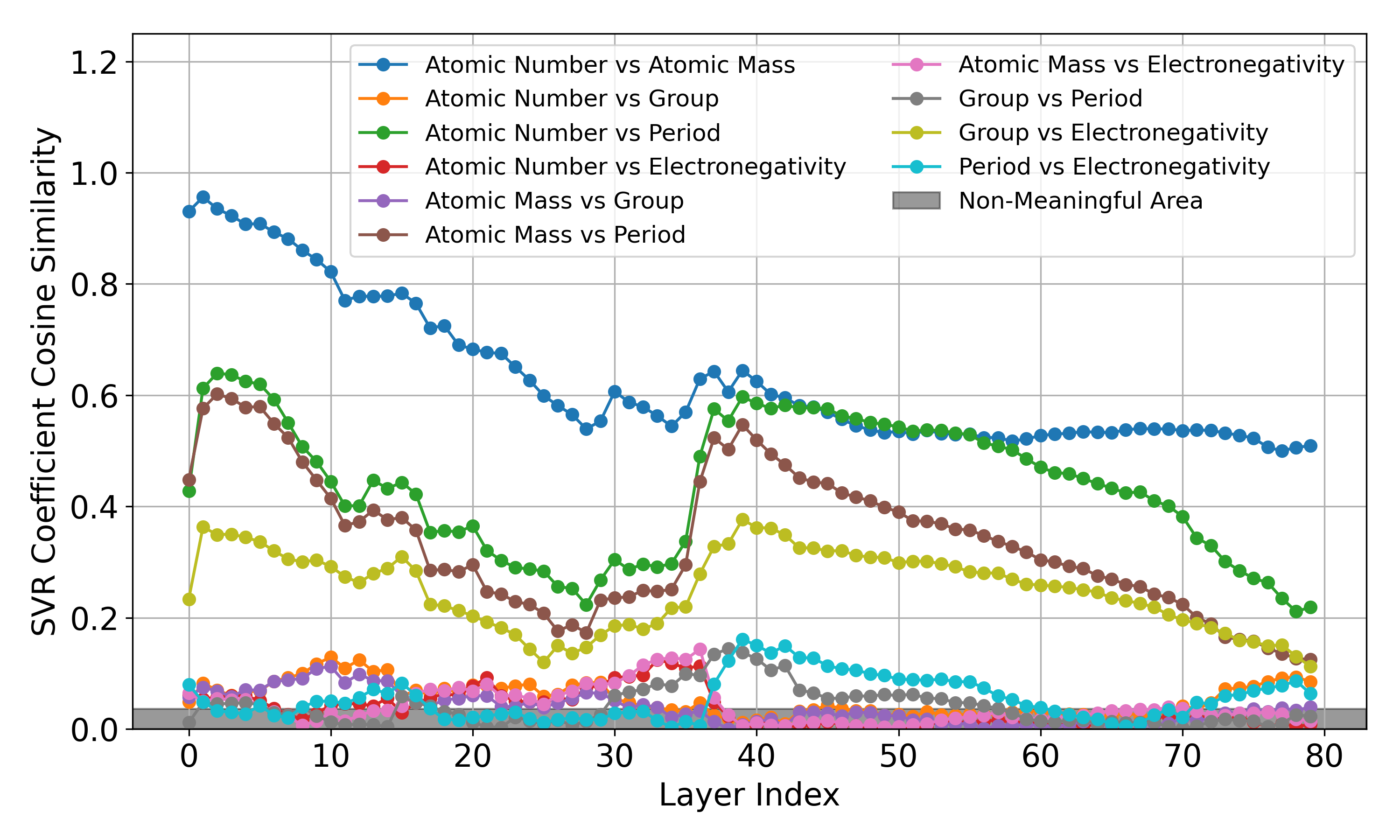}
    \caption{Cosine similarity between weight vectors of linear probes for attribute pairs across layers in Meta-Llama-3.1-70B. The shaded area (99.9\% CI) represents unrelated boundaries.}
    \label{fig:similarity}
\end{figure}

Fig. \ref{fig:similarity} illustrates the cosine similarity across 80 layers of Meta-Llama-3.1-70B. Notably, in high-dimensional spaces, random vector pairs typically approach orthogonality due to the `blessing of dimensionality'. To illustrate this, we randomly sampled vector pairs in an 8129-dimensional space (the residual stream vector size of Meta-Llama-3.1-70B) and calculated their cosine similarity, with the 99.9\% confidence interval (CI) shown in gray. Cosine similarity outside this interval indicates meaningful relationships between attributes. See Appendix \ref{app:blessing} for more details.

In the early layers, high similarity reflects token-level processing rather than semantic understanding. As layers deepen, similarity decreases as the model begins capturing semantics. In the intermediate layers, similarity rises, indicating shared representation of correlated attributes. Finally, in the later layers, similarity drops again as the model separates features for refined decision-making.

\subsection{Blessing of dimensionality}
\label{app:blessing}

When the dimensionality is very high, the most of random vector pairs approach orthogonality. We illustrate this by sampling pairs of vectors in an 8129-dimensional space (corresponding to the residual stream vector dimension of Meta-Llama-3.1-70B) and computing their cosine similarities. The 99.9\% confidence interval (CI) provides an estimate of the expected cosine similarity range at each dimensionality:  

\[
\text{CI}_{99.9\%} = \left( \mu - z \frac{\sigma}{\sqrt{n}}, \mu + z \frac{\sigma}{\sqrt{n}} \right)
\]

where \( \mu \) is the sample mean, \( \sigma \) is the sample standard deviation, \( n \) is the number of sampled pairs, and \( z \approx 3.29 \) for a 99.9\% confidence level.  

\begin{figure*}[h]
\centering
    \includegraphics[width=0.4\linewidth]{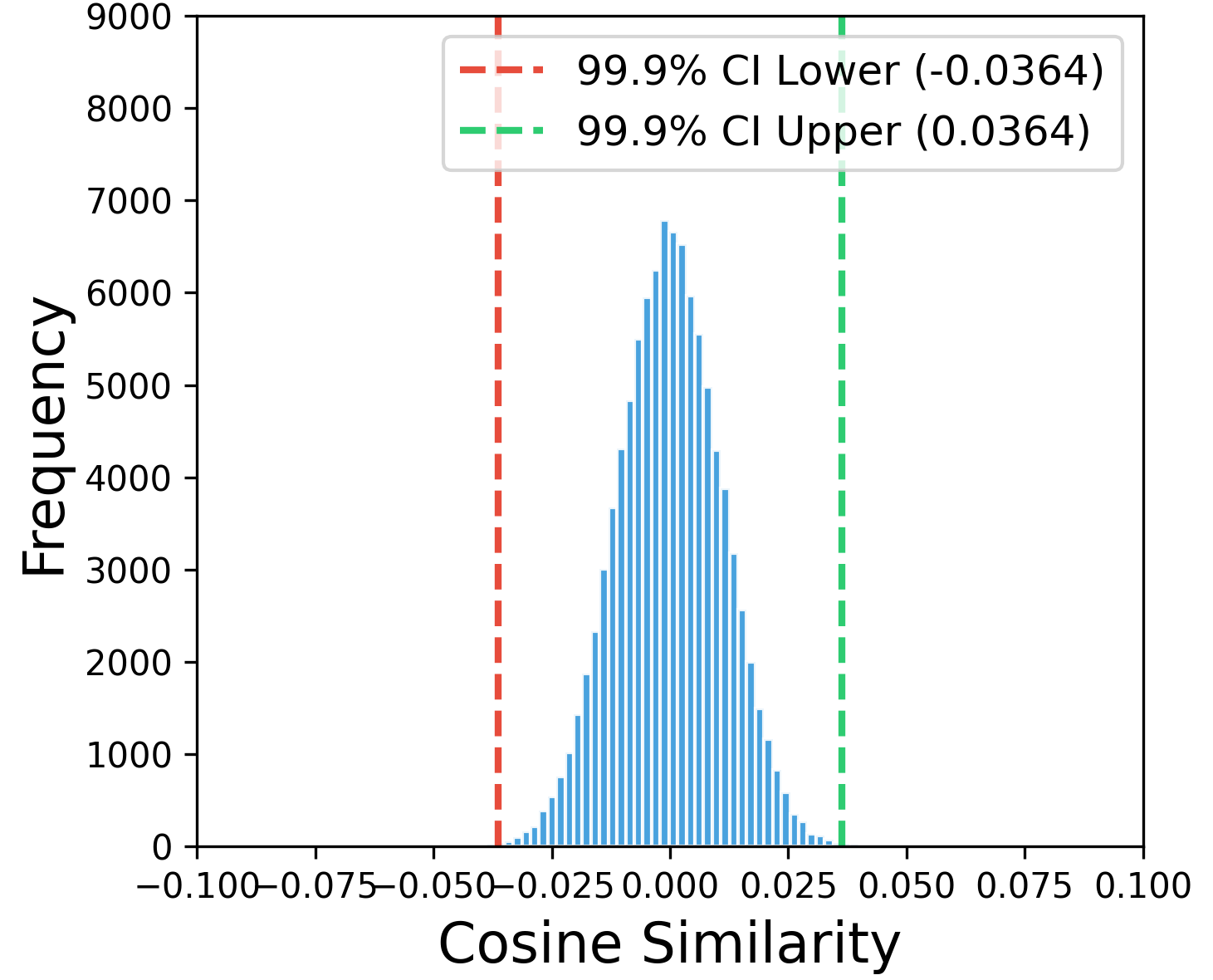}
     \caption{Cosine similarity distribution of random vector pairs in an 8129-dimensional space, with a 99.9\% confidence interval ($-0.0364$, $0.0364$) shown by the dashed lines.}

    \label{fig:CI}
\end{figure*}

The specific distribution is shown in Fig. \ref{fig:similarity}, where the confidence interval is extremely narrow (\(\pm 0.036\)), indicating that random vector pairs exhibit highly consistent cosine similarities. This suggests that the learned weights of the linear probe across different feature pairs are effectively uncorrelated, exhibiting only random alignment. In Fig. \ref{fig:distance}, the shaded region represents the 99.9\% confidence interval for the cosine similarities of high-dimensional random vectors, further supporting this observation.

\section{Linear probing detailed results}
\label{app:linear_probing}

In Sec.\ref{sub:probing}, we applied linear probing to train linear-kernel Support Vector Regression (SVR) models for each layer \( l \) and attribute \( A_j \), using the residual stream dataset \( H_j(l) \). The ground truth values \( y_{i,j} \) correspond to the attribute values of each element \( X_i \).

\subsection{Why \( R^2 \) cannot reach 1}
\label{app:not_1}
Even with a perfectly trained model and a sufficiently large dataset, achieving an \( R^2 \) of 1 in linear probing is impossible. The model’s output token '1' indicates that the residual stream at the last token position in the final layer leads to the highest logit for token 1’s ID after the final linear transformation. However, do token residual stream in the final layer of numbers exhibit a perfect linear relationship with their real numerical values? Token embeddings are learned representations that capture semantic relationships between tokens, but they are not guaranteed to align linearly with numerical values.  

To further investigate this, we fit a linear model to map approximated numerical token representations in the last layer to their actual values. Specifically, we extract token IDs for numbers 1–50 from the tokenizer, multiply them by the pseudoinverse of the vocabulary projection matrix \( \mathbf{W}_{\text{vocab}}^+ \), and obtain their corresponding vector representations in the hidden space of the last layer:

\[
\mathbf{h}_i = \mathbf{W}_{\text{vocab}}^+ \cdot \mathbf{t}_i
\]

where \( \mathbf{t}_i \) is the encoded token ID for the number \( i \), and \( \mathbf{h}_i \) represents its corresponding hidden space representation.

We then fit a linear regression model to map these representations to their true numerical values. The linear correlation turned out to be quite strong, with an \( R^2 \) of 0.98. However, this is not 1—possibly because the embedding space is not perfectly linearly aligned with numerical values, or because it is influenced by semantic noise, or simply due to limitations in the fitting method.  

For LLMs, even if the logits were identical to the embeddings \( \mathbf{h}_i \) (which is theoretically impossible—at best, they can only approximate them), the \( R^2 \) would still be limited to 0.98. Therefore, it is unsurprising that linear probing does not achieve an \( R^2 \) of 1.  

\subsection{Confusion matrix}
\label{app:confusion}

\begin{figure}[h]
\centering
    \includegraphics[width=0.4\linewidth]{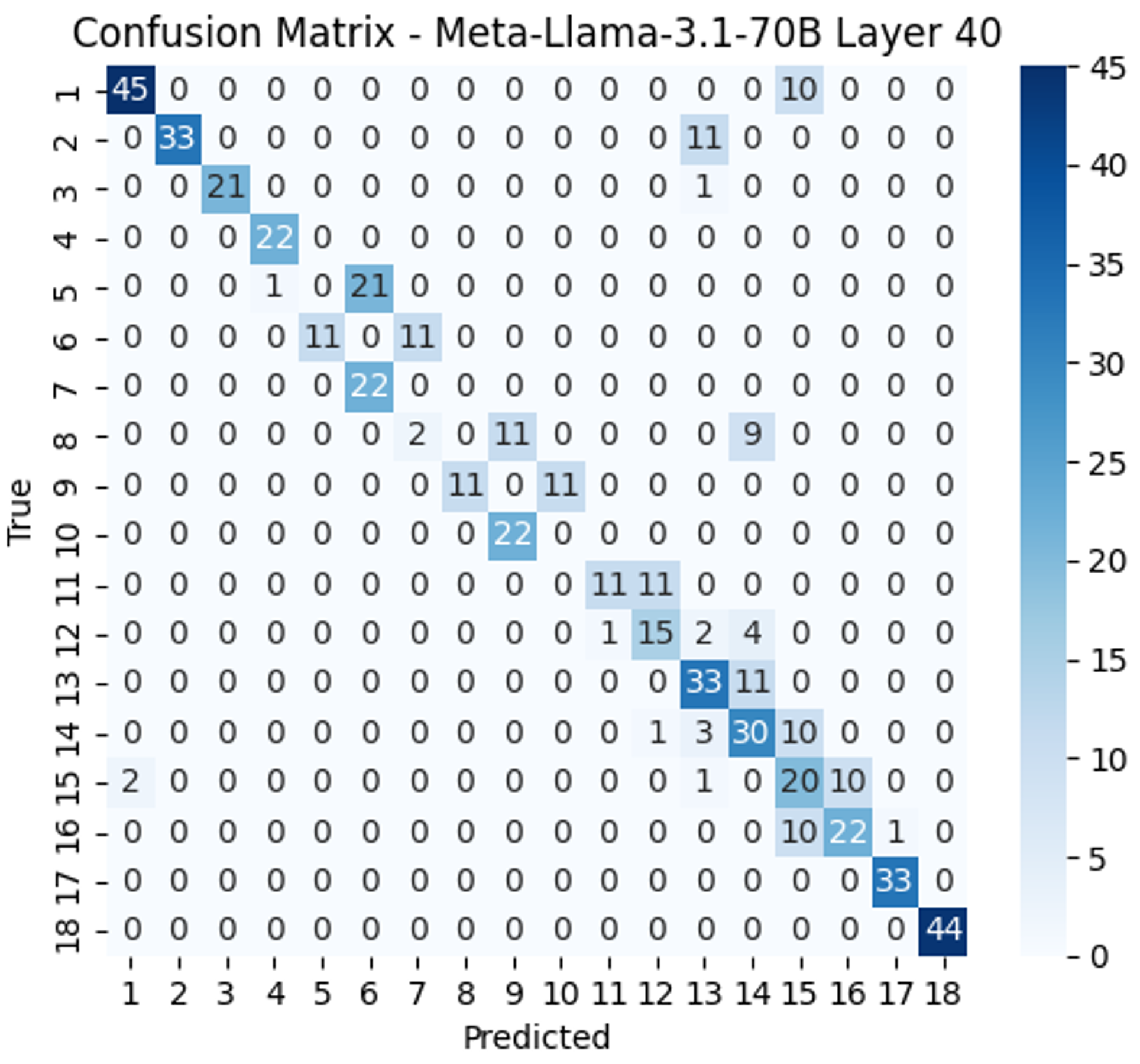}
    \caption{Confusion Matrix on categorical linear probing of attribute ‘Group’ on the middle layer. }
    \label{fig:confusion}
\end{figure}

\subsection{Question prompts}
\label{app: question_vs}

We applied the Mann–Kendall test to $\Delta R^2$ in the upper 50\% of layers for each model–attribute pair (15 in total).  
Benjamini–Hochberg adjustment (FDR = 0.05) retained significance in 12 pairs, with τ ranging from 0.36 to 0.94 (median 0.79; Table \ref{tab:mk_results}).  

\begin{figure*}[h]
\centering
    \includegraphics[width=1\linewidth]{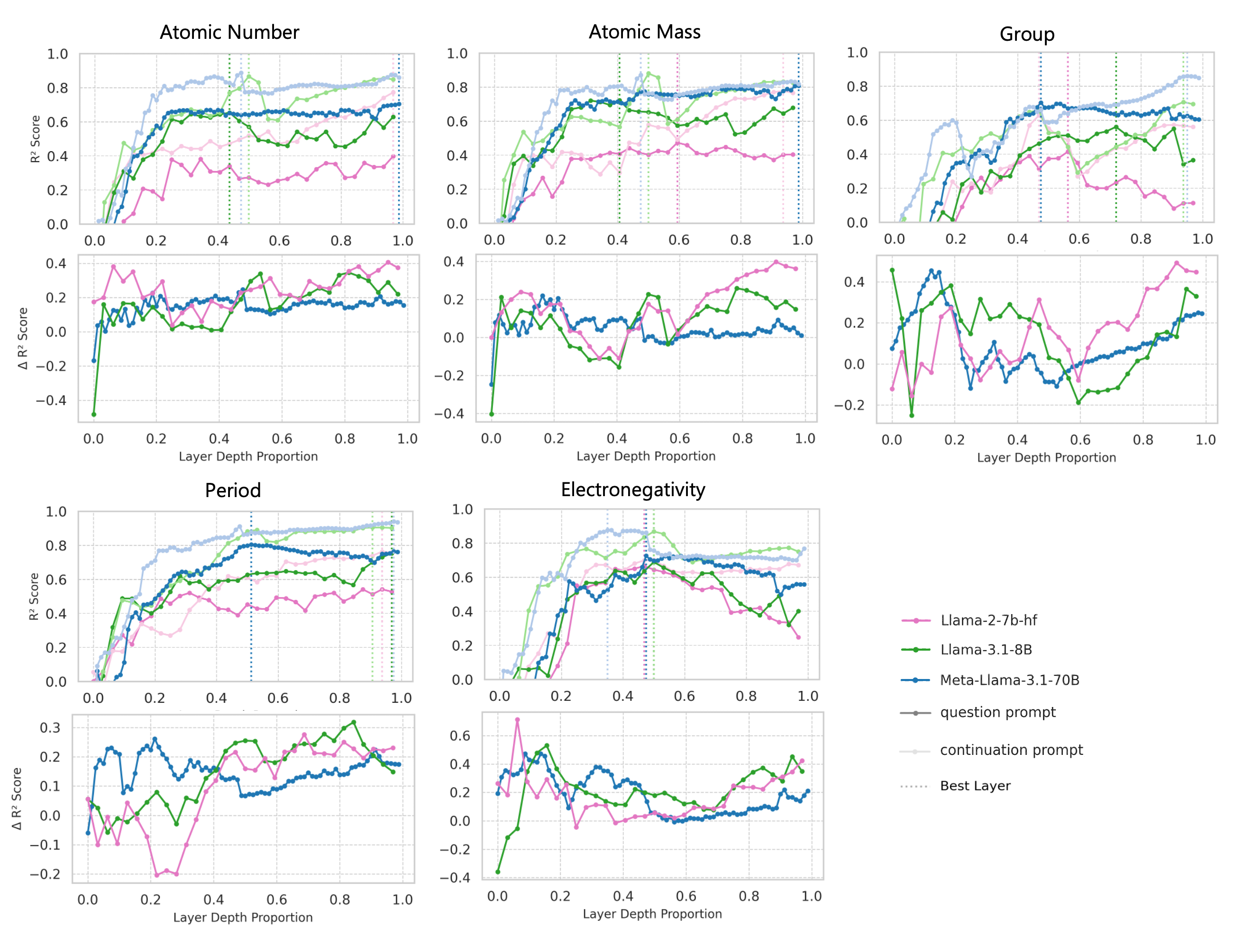}
    \caption{R² scores for linear probes trained on target properties and evaluated on representations from question and continuation prompts. }
    \label{fig:question_vs_con}
\end{figure*}

\begin{table}[h]
\centering
\caption{Mann--Kendall trend test on $\Delta R^2$ (difference between continuation and question prompt R²) within depth layer depths 0.5--1.0.  $\tau$: Kendall's tau (trend strength).  (*) indicates significance after Benjamini--Hochberg FDR correction ($\alpha=0.05$).}
\label{tab:mk_results}
\begin{tabular}{llrr}
\toprule
Model & Attribute & $\tau$ & $p$ \\
\midrule
\multirow{5}{*}{Meta--Llama--3.1--70B}
  & Group             & 0.95 & $<0.001$* \\
  & Period            & 0.84 & $<0.001$* \\
  & Electronegativity & 0.73 & $<0.001$* \\
  & Atomic Mass       & 0.53 & $<0.001$* \\
  & Atomic Number     & 0.41 & $<0.001$* \\
\addlinespace
\multirow{5}{*}{Llama--2--7b--hf}
  & Electronegativity & 0.80 & $<0.001$* \\
  & Atomic Mass       & 0.80 & $<0.001$* \\
  & Group             & 0.75 & $<0.001$* \\
  & Atomic Number     & 0.55 & $0.003$* \\
  & Period            & 0.40 & $0.034$* \\
\addlinespace
\multirow{5}{*}{Llama--3.1--8B}
  & Group             & 0.57 & $0.003$* \\
  & Electronegativity & 0.52 & $0.006$* \\
  & Atomic Number     & 0.20 & $0.300$  \\
  & Atomic Mass       & 0.20 & $0.300$  \\
  & Period            & -0.05 & $0.822$  \\
\bottomrule
\end{tabular}
\end{table}

\clearpage
\subsection{`Non-matching' and `no mention' prompts}
\label{app: matching_vs}

\subsubsection{Non-matching attribute pairs}
\label{par:pair_choosing}
To contrast explicit and implicit attribute cues, we selected a set of
\emph{non-matching} attribute pairs with minimal direct statistical dependency.
Each pair was chosen to lack strong linear or monotonic relationships, and we verified
on our 50-element subset that all selected pairs satisfy
$|{\rm Pearson}\,r|<0.30$, $|{\rm Spearman}\,\rho|<0.30$,
and $R^{2}<0.15$ (Table~\ref{tab:nonmatching_corr}).
These thresholds ensure that no pair can be trivially recovered through
simple linear or monotonic mappings.

\begin{table}[h]
\centering
\caption{Linear and monotonic correlations for non-matching attribute pairs (50-element subset).}
\label{tab:nonmatching_corr}
\begin{tabular}{lrrr}
\toprule
        Pair &  |Pearson r| &  |Spearman ρ| &  Linear R² \\
\midrule
     Group–Atomic Number &     0.044 &      0.070 &      0.002 \\
Group–Period &     0.255 &      0.300 &      0.065 \\
  Group–Mass &     0.037 &      0.066 &      0.001 \\
        Electronegativity–Atomic Number &     0.154 &      0.038 &      0.024 \\
     Electronegativity–Mass &     0.147 &      0.039 &      0.022 \\
\bottomrule
\end{tabular}
\end{table}

\subsubsection{Detailed results}

\begin{figure}[h]
\centering
    \includegraphics[width=1\linewidth]{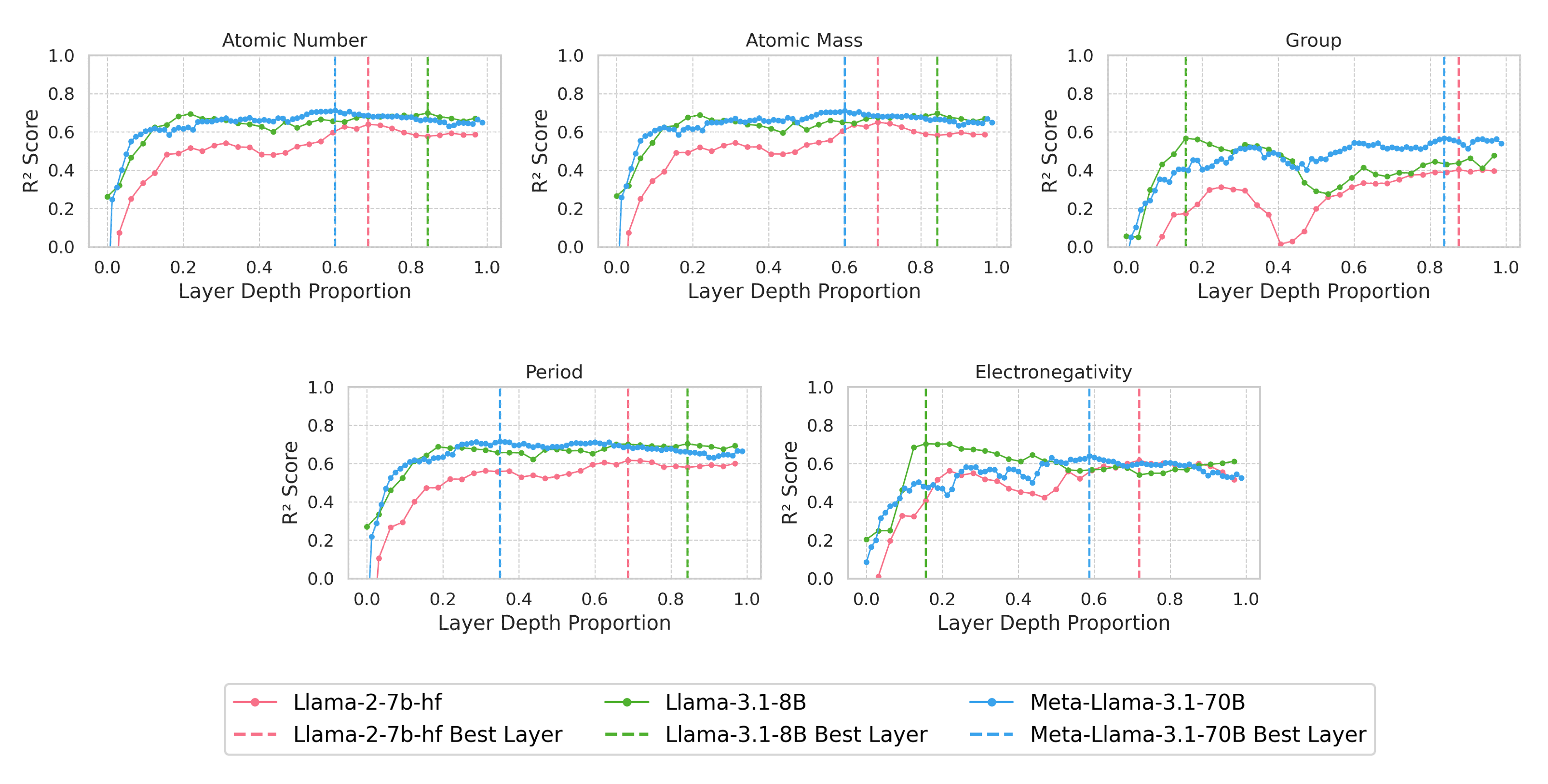}
    \caption{R² score trends for `no mention' cases. Regression linear probing on the element token residual stream with 5-fold cross-validation was performed on residual streams, and R2 scores on the test set are shown for each attribute.}
    \label{fig:regression_ele}
\end{figure}

\begin{figure*}[h]
\centering
    \includegraphics[width=1\linewidth]{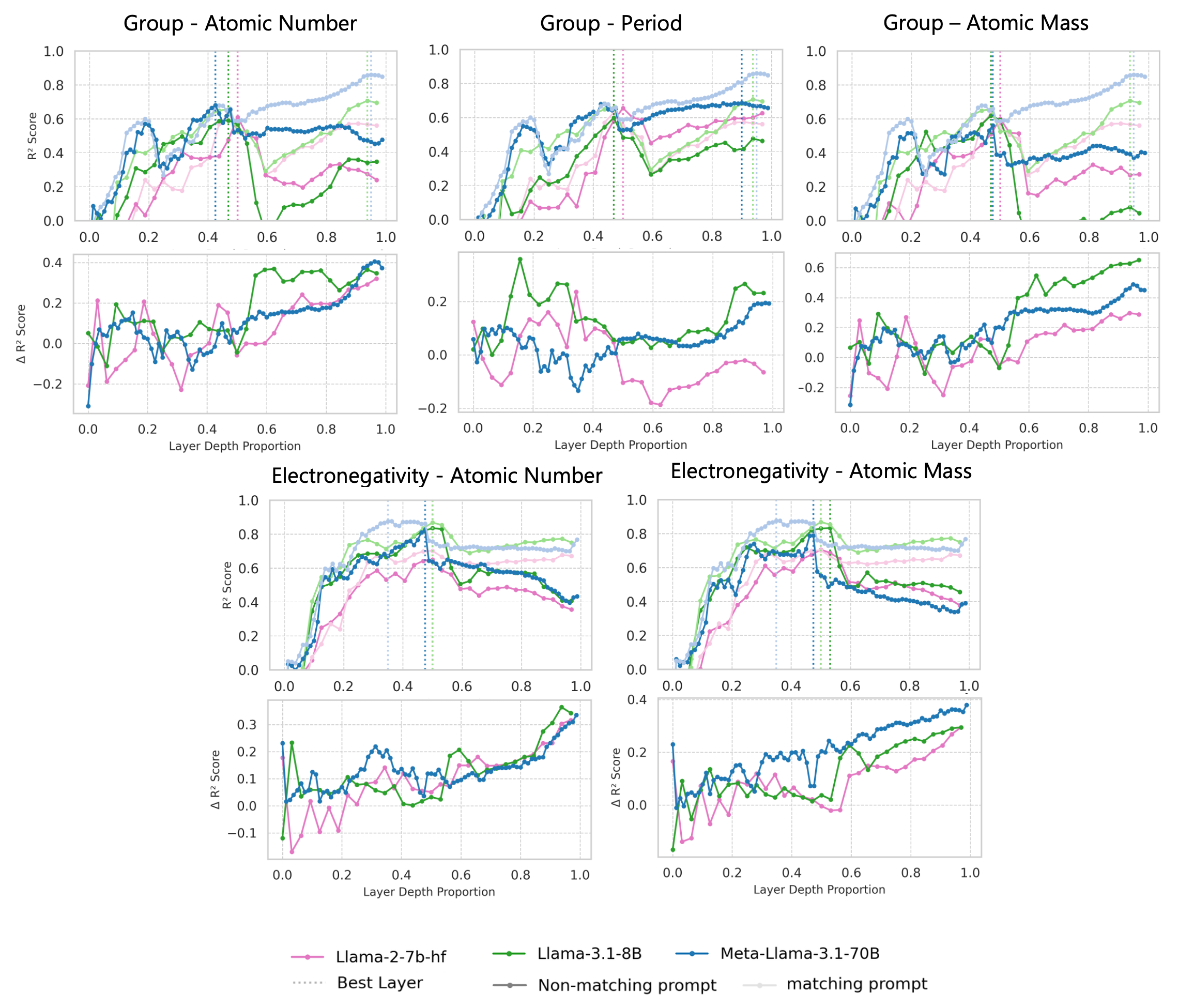}
    \caption{R² and \(\Delta R^2\) scores for linear probes trained on target properties and evaluated on representations from matching and non-matching prompts. \(\Delta R²\) is defined as R²(matching prompt) minus R²(non-matching prompt).}
    \label{fig:matching_detail}
\end{figure*}

\begin{figure*}[h]
\centering
    \includegraphics[width=0.6\linewidth]{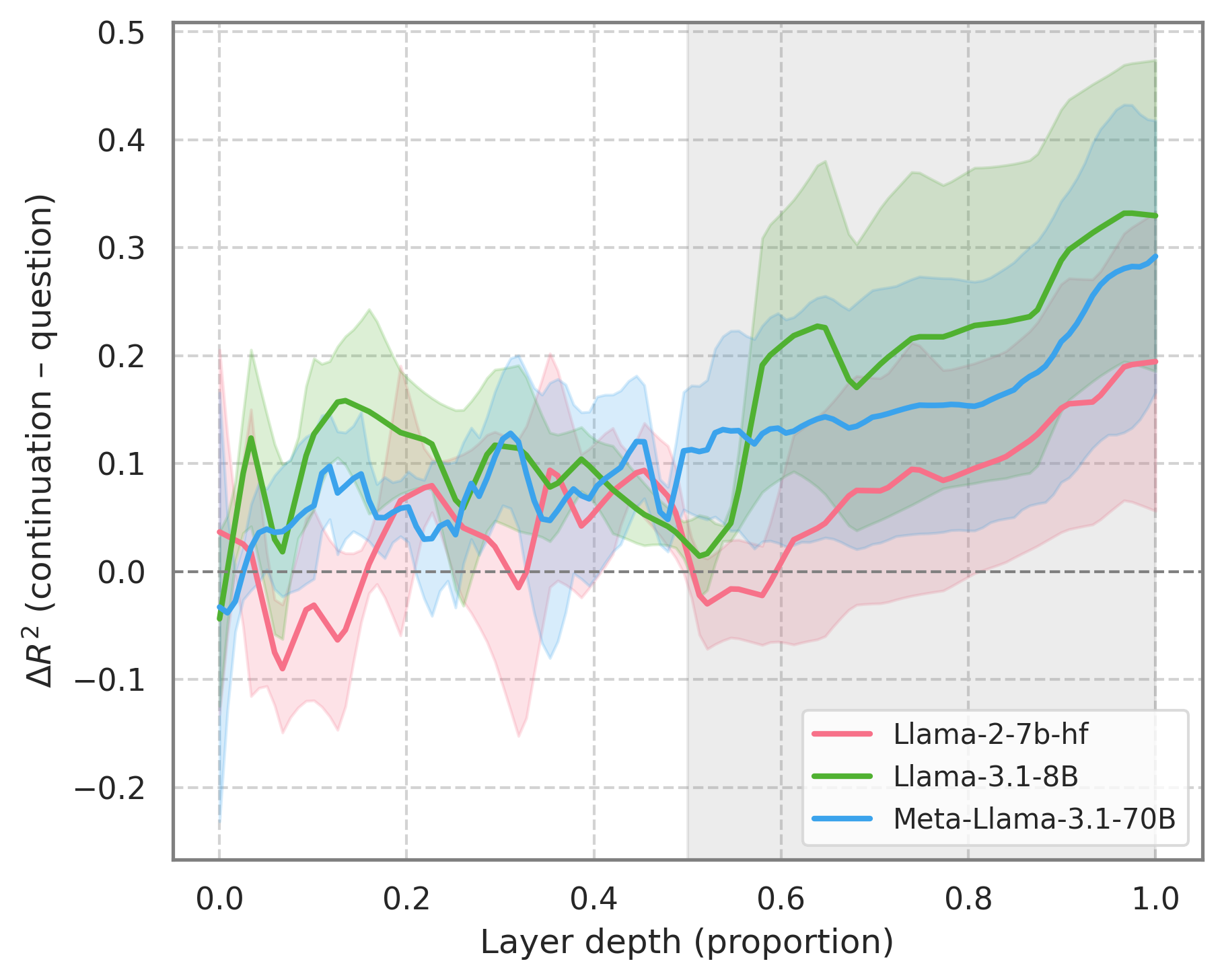}
    \caption{Average \(\Delta R^2\) across five attributes, with with 95\% confidence interval shaded. \(\Delta R^2 = R^2_{\text{match}} - R^2_{\text{non-match}}\).}
    \label{fig:delta_nonmatching}
\end{figure*}

\begin{table}[h]
\centering
\caption{Mann--Kendall trend test on $\Delta R^2$ (difference between matching and non-matching prompt R²) within layer depths 0.6--1.0. $\tau$: Kendall’s tau (trend strength). “*” indicates significance after Benjamini–Hochberg FDR correction ($\alpha=0.05$).}
\label{tab:mk_results}
\begin{tabular}{llrr}
\toprule
Model & Attribute Pair & $\tau$ & $p$ \\
\midrule
\multirow{6}{*}{Meta–Llama–3.1–70B}
  & Electronegativity – Atomic Number & $0.923$ & $<0.001$* \\
  & Group – Atomic Number            & $0.911$ & $<0.001$* \\
  & Electronegativity – Atomic Mass  & $0.895$ & $<0.001$* \\
  & Group – Period                   & $0.625$ & $<0.001$* \\
  & Group – Atomic Mass              & $0.452$ & $<0.001$* \\
\addlinespace
\multirow{6}{*}{Llama–2–7b–hf}
  & Group – Atomic Number            & $0.872$ & $<0.001$* \\
  & Group – Atomic Mass              & $0.846$ & $<0.001$* \\
  & Electronegativity – Atomic Mass  & $0.795$ & $<0.001$* \\
  & Group – Period                   & $0.769$ & $<0.001$* \\
  & Electronegativity – Atomic Number & $0.769$ & $<0.001$* \\
\addlinespace
\multirow{6}{*}{Llama–3.1–8B}
  & Electronegativity – Atomic Mass  & $0.821$ & $<0.001$* \\
  & Group – Period                   & $0.769$ & $<0.001$* \\
  & Group – Atomic Mass              & $0.744$ & $<0.001$* \\
  & Electronegativity – Atomic Number & $0.641$ & $0.003$* \\
  & Group – Atomic Number            & $-0.154$ & $0.502$   \\
\bottomrule
\end{tabular}
\end{table}

\clearpage
\subsection{Detailed results of the best layer}
\label{app:best_layer}
\begin{figure*}[h]
\centering
    \includegraphics[width=1\linewidth]{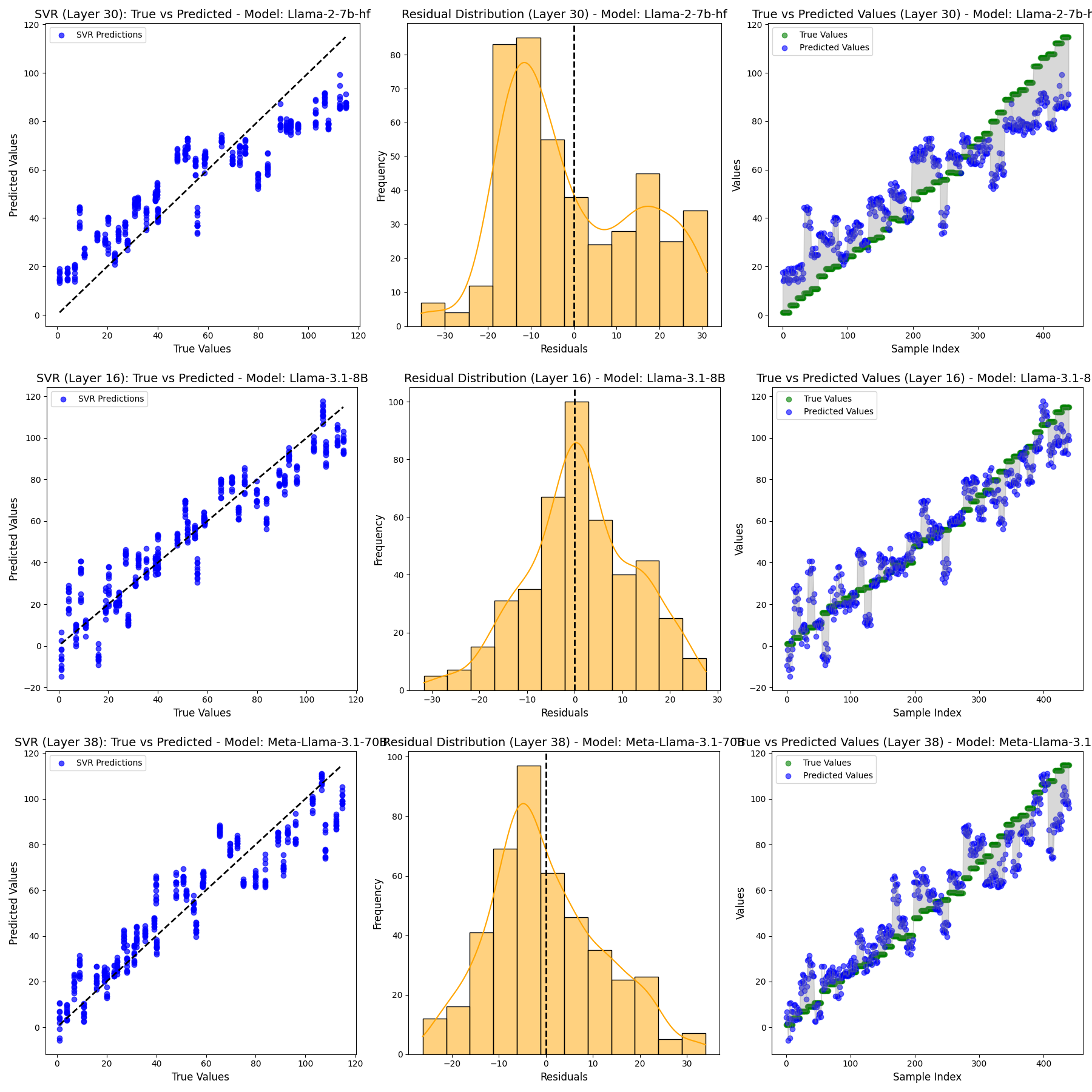}
    \caption{Evaluation of SVR performance for Layer \( \text{best\_layer} \) on the atomic mass. The left plot shows true vs. predicted values with alignment to the diagonal indicating accuracy. The center plot displays residuals, highlighting error distribution centered around zero. The right plot visualizes true and predicted values across samples, with shaded areas representing error magnitudes.}
    \label{fig:mass}
\end{figure*}

\begin{figure*}[h]
\centering
    \includegraphics[width=1\linewidth]{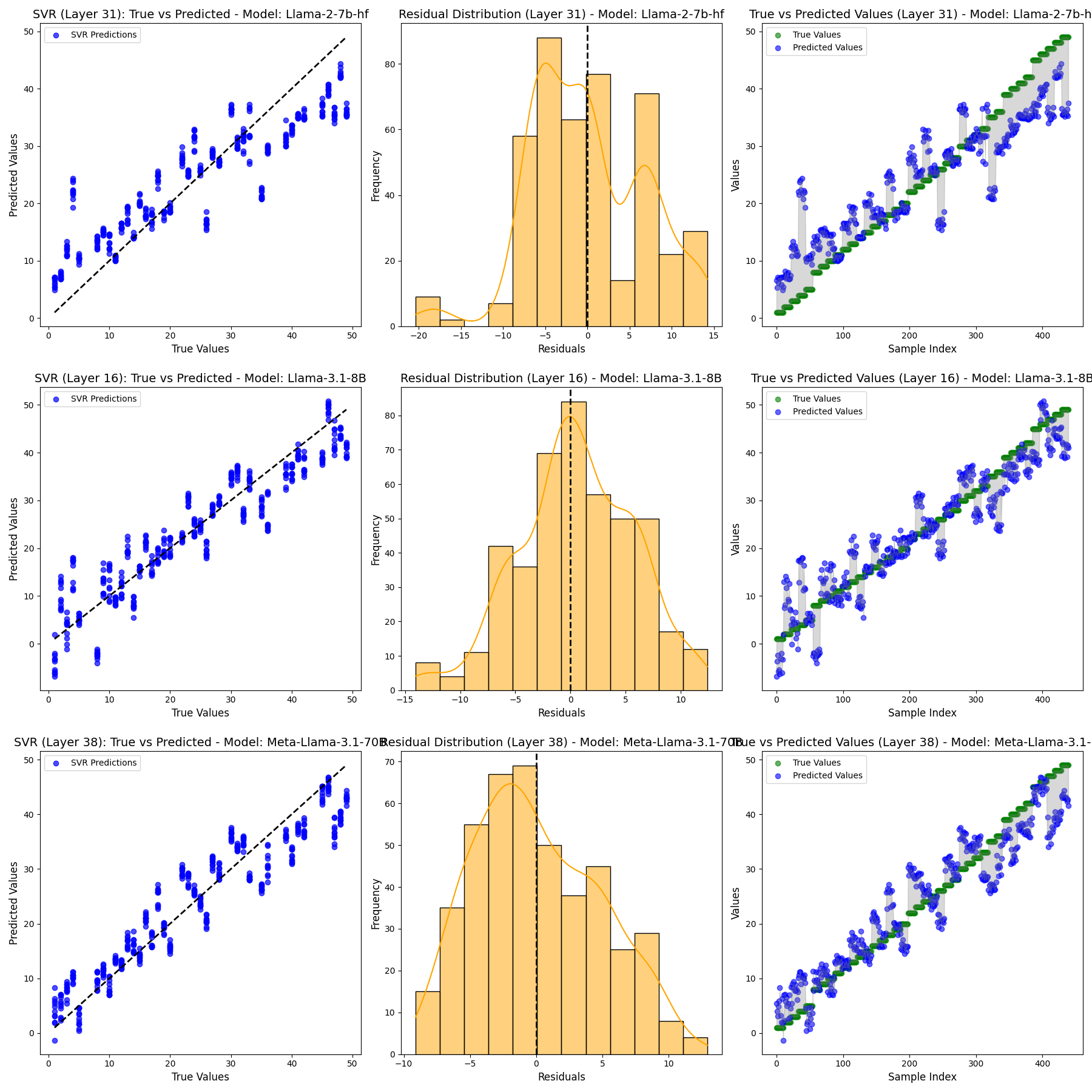}
    \caption{Evaluation of SVR performance for Layer \( \text{best\_layer} \) on the atomic number. The left plot shows true vs. predicted values with alignment to the diagonal indicating accuracy. The center plot displays residuals, highlighting error distribution centered around zero. The right plot visualizes true and predicted values across samples, with shaded areas representing error magnitudes.}
    \label{fig:number}
\end{figure*}

\begin{figure*}[h]
\centering
    \includegraphics[width=1\linewidth]{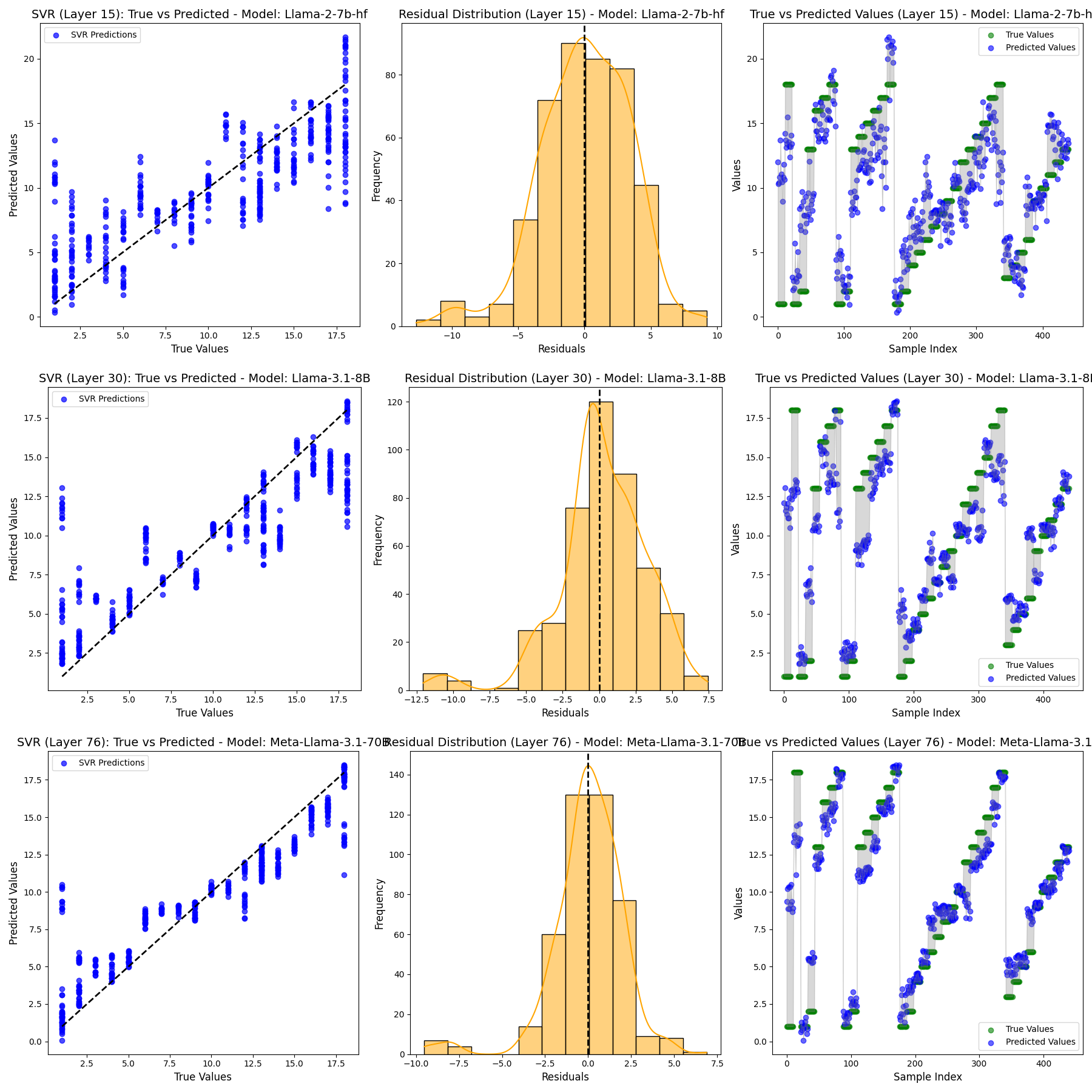}
    \caption{Evaluation of SVR performance for Layer \( \text{best\_layer} \) on the group. The left plot shows true vs. predicted values with alignment to the diagonal indicating accuracy. The center plot displays residuals, highlighting error distribution centered around zero. The right plot visualizes true and predicted values across samples, with shaded areas representing error magnitudes.}
    \label{fig:group}
\end{figure*}

\begin{figure*}[h]
\centering
    \includegraphics[width=1\linewidth]{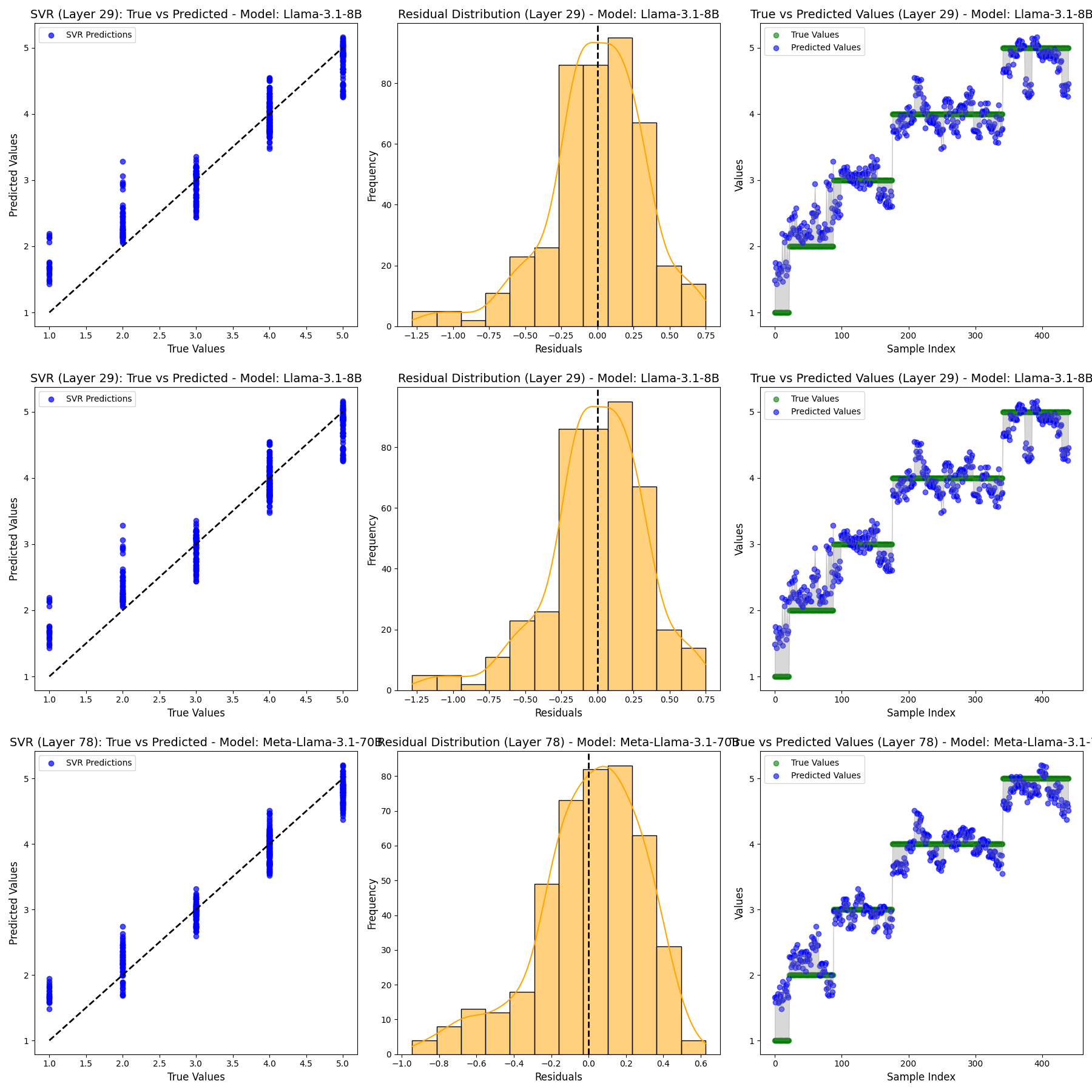}
    \caption{Evaluation of SVR performance for Layer \( \text{best\_layer} \) on the period. The left plot shows true vs. predicted values with alignment to the diagonal indicating accuracy. The center plot displays residuals, highlighting error distribution centered around zero. The right plot visualizes true and predicted values across samples, with shaded areas representing error magnitudes.}
    \label{fig:period}
\end{figure*}

\begin{figure*}[h]
\centering
    \includegraphics[width=1\linewidth]{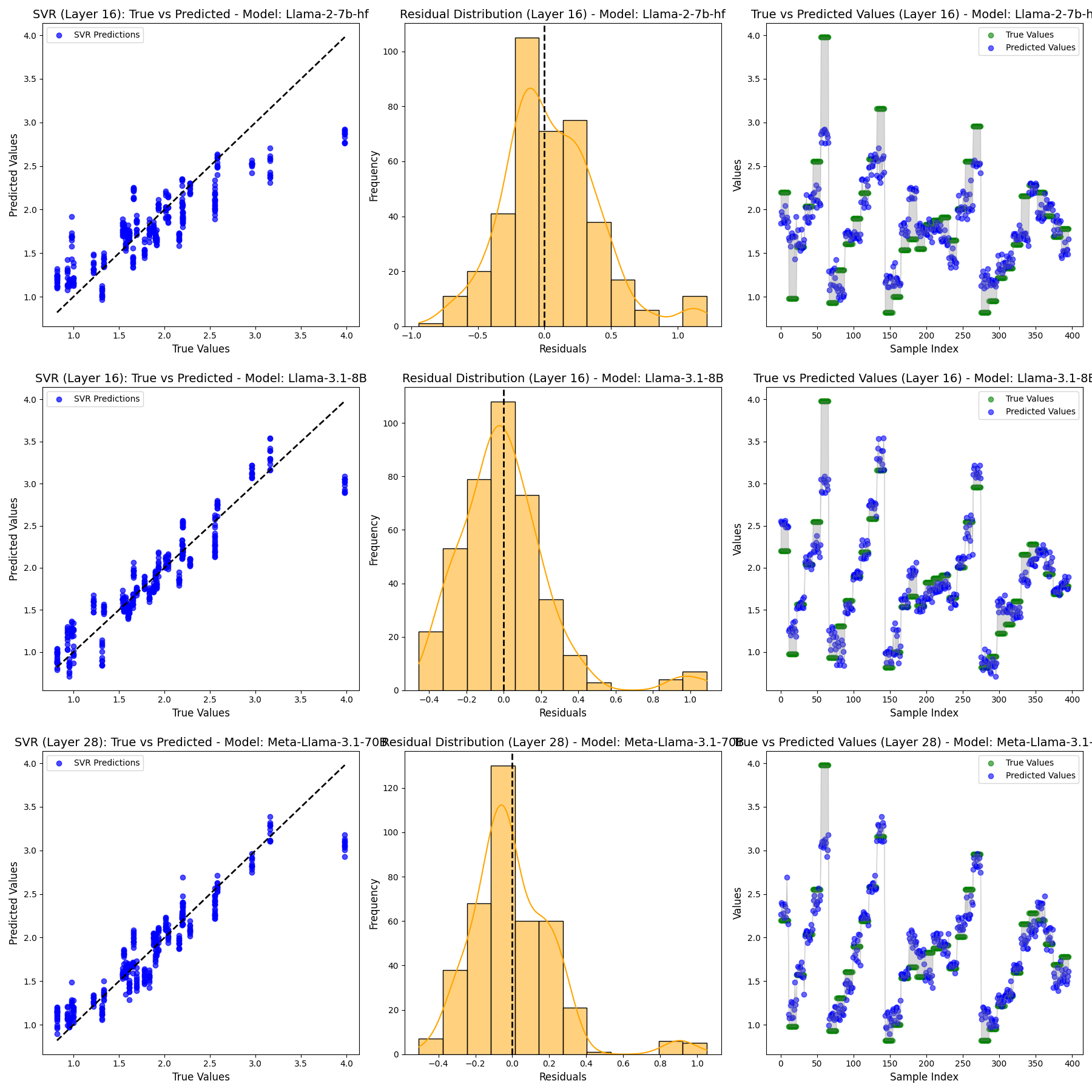}
    \caption{Evaluation of SVR performance for Layer \( \text{best\_layer} \) on the electronegativity. The left plot shows true vs. predicted values with alignment to the diagonal indicating accuracy. The center plot displays residuals, highlighting error distribution centered around zero. The right plot visualizes true and predicted values across samples, with shaded areas representing error magnitudes.}
    \label{fig:elec}
\end{figure*}

In the main text Sec. \ref{sub:probing}, we used SVR for linear regression probing. Figures~\ref{fig:mass}, \ref{fig:number}, \ref{fig:group}, \ref{fig:period}, and \ref{fig:elec} present the detailed \(R^2\) performance of the best layer for each attribute—atomic number, atomic mass, electronegativity, period, and group—across three different models.

